\documentclass[journal]{IEEEtran}

\title{From Hand-Crafted to Deep Learning-based Cancer Radiomics: Challenges and Opportunities\thanks{This work was partially supported by the Fonds de Recherche du Qu\'ebec Nature et Technologies (FRQNT) Grant 206591 EQ. Corresponding Author is Arash Mohammadi, email: arash.mohammadi@concordia.ca}}
\author{Parnian Afshar$\dag$,~\IEEEmembership{Student Member,~IEEE}, Arash Mohammadi$\dag$,~\IEEEmembership{Senior Member,~IEEE}, Konstantinos N. Plataniotis$\ddag$,~\IEEEmembership{Fellow,~IEEE}, Anastasia Oikonomou$^{*}$, and Habib Benali$\top$,~\IEEEmembership{Member,~IEEE}\\\vspace{.1in}
$\dag$Concordia Institute for Information Systems Engineering (CIISE),  Concordia University, Montreal,~Canada \\\vspace{.05in}
$\ddag$ Department of Electrical and Computer Engineering, University of Toronto, Toronto, Canada\\\vspace{.05in}
$^{*}$Department of Medical Imaging, Sunnybrook Health Sciences Centre, University of Toronto, Toronto,~Canada\\\vspace{.05in}
$\top$PERFORM Centre, Electrical and Computer Engineering Department,  Concordia University, Montreal,~Canada}
\usepackage[dvipdfmx]{graphicx}
\usepackage{epsfig}
\usepackage{amsmath}
\usepackage{multirow}
\usepackage{setspace}

\usepackage{multicol}
\usepackage{cuted,lipsum}
\usepackage{amsthm}
\usepackage{amssymb }
\usepackage{amsmath}
\usepackage{bm}
\usepackage{cite}
\usepackage{float}
\usepackage{mathrsfs}
\usepackage{amsfonts}
\usepackage{algorithm, algpseudocode}
\usepackage{subfigure}
\usepackage[usenames]{color}
\usepackage[dvipsnames]{xcolor}
\usepackage{tcolorbox}
\usepackage{lipsum}
\usepackage{booktabs,caption}
\usepackage{tabularx}
\usepackage{multirow}
\usepackage{bm}
\usepackage{tabu}
\usepackage{xcolor,colortbl}
\usepackage{lscape}
\usepackage{nomencl}

\newcolumntype{Y}{>{\centering\arraybackslash}X}

\makenomenclature

\definecolor{Gray}{gray}{0.85}
\definecolor{LightCyan}{rgb}{0.6,0.9,1}

\newcolumntype{s}{>{\hsize=.7\hsize}Y}

\usepackage{tikz}

\def\DLR{\text{DLR}} 
\def\HCR{\text{HCR}} 
\def\i{^{(i)}}
\def\NS{N_{s}}
\def\W{\bm{W}}
\def\b{\bm{b}}
\def\c{\bm{c}}
\def\u{\bm{u}}
\def\s{\bm{s}}

\def\x{\bm{x}}

\def\f{\bm{f}}
\begin{document}

\date{\today}
\maketitle
\thispagestyle{empty}

\begin{abstract}
Recent advancements in signal processing and machine learning coupled with developments of electronic medical record keeping in hospitals and the availability of extensive set of medical images through internal/external communication systems, have resulted in a recent surge of significant interest in ``Radiomics''. Radiomics is an emerging and relatively new research field, which refers to extracting semi-quantitative and/or quantitative features from medical images with the goal of developing predictive and/or prognostic  models, and is expected to become a critical component for integration of image-derived information for personalized treatment in the near future. The conventional Radiomics workflow is typically based on extracting pre-designed features (also referred to as hand-crafted or engineered features)  from a segmented region of interest. Nevertheless, recent advancements in deep learning have caused trends towards deep learning-based Radiomics (also referred to as discovery Radiomics). Considering the advantages of these two approaches, there are also hybrid solutions developed to exploit the potentials of multiple data sources. Considering the variety of approaches to Radiomics, further improvements require a comprehensive and integrated sketch, which is the goal of this article. This manuscript provides a unique interdisciplinary perspective on Radiomics by discussing state-of-the-art signal processing solutions in the context of Radiomics.
\end{abstract}
\nomenclature{HCR}{Hand-Crafted Radiomics}
\nomenclature{DLR}{Deep Learning-based Radiomics}
\nomenclature{CT}{Computed Tomography}
\nomenclature{MRI}{Magnetic Resonance Imaging}
\nomenclature{PET}{Positron Emission Tomography}
\nomenclature{ROI}{Region on Interest}
\nomenclature{VOI}{Volume of Interest}
\nomenclature{GLCM}{Gray Level Co-occurrence Matrix}
\nomenclature{GLRLM}{Gray Level Run Length Matrix}
\nomenclature{NGTDM}{Neighborhood Gray Tone Difference Matrix}
\nomenclature{LoG}{Laplacian of Gaussian}
\nomenclature{PCA}{Principle Component Analysis}
\nomenclature{ICA}{Independent Component Analysis}
\nomenclature{ZV}{Zero Variance}
\nomenclature{FSCR}{Fisher Score}
\nomenclature{MIFS}{Mutual Information Feature Selection}
\nomenclature{MRMR}{Minimum Redundancy Maximum Relevancy}
\nomenclature{MDS}{Multidimensional Scaling}
\nomenclature{ISOMAP}{Isometric Mapping}
\nomenclature{LLE}{Locally Linear Embedding}
\nomenclature{SVM}{Support Vector Machine}
\nomenclature{NN}{Neural Network}
\nomenclature{KMS}{Kaplan-Meier Survival}
\nomenclature{PHM}{Cox Proportional Hazard Model}
\nomenclature{GLZLM}{Generalized Linear Model}
\nomenclature{GLZLM}{Grey-Level Zone Length Matrix}
\nomenclature{NB}{Naive Bayes}
\nomenclature{KNN}{K-Nearest Neighborhood}
\nomenclature{MDA}{Mixture Discriminant Analysis}
\nomenclature{PLS}{Partial Least Squares}
\nomenclature{FP}{False Positive}
\nomenclature{FN}{False Negative}
\nomenclature{ROC}{Receiver Operating Characteristic}
\nomenclature{ICC}{Intra-Class Correlation Coefficient}
\nomenclature{ANOVA}{Analysis of Variance}
\nomenclature{SRC}{Spearman's Rank Correlation Coefficient}
\nomenclature{GSEA}{Gene Set Enrichment Analysis}
\nomenclature{CNN}{Convolutional Neural Network}
\nomenclature{RNN}{Recurrent Neural Network}
\nomenclature{DAE}{Denoising Auto-Encoder}
\nomenclature{CAE}{Convolutional Auto-Encoder}
\nomenclature{DBN}{Deep Belief Network}
\nomenclature{RBM}{Restricted Boltzmann Machine}
\nomenclature{DBM}{Deep Boltzmann Machine}
\nomenclature{CapsNet}{Capsule Network}
\nomenclature{LSTM}{Long-Short Term Memory}
\nomenclature{AUC}{Area Under Curve}
\nomenclature{T-SNE}{T-distributed stochastic neighbor embedding}
\nomenclature{HoG}{Histogram of Gradients}
\nomenclature{GPU}{Graphic Processing Unit}
\nomenclature{MF}{Minkowski Functional}

\section{Introduction} \label{sec:Introduction}
The volume, variety, and velocity of medical imaging data generated for medical diagnosis are exploding. Generally speaking, medical diagnosis refers to determining the source and etiology of a medical condition. Diagnosis is typically reached  by means of several medical tests, among them biopsy and diagnostic imaging,  in case of suspected cancer. Although biopsy can be very informative, it is invasive and by being focal, may not represent the heterogeneity of the entire tumor, which is crucial in cancer prognosis and treatment. In contrast to biopsy, diagnostic imaging is not invasive and can provide information on tumor's overall shape, growth over time, and heterogeneity, making it an attractive and favored alternative to biopsy. Interpretation of such a large amount of diagnostic images, however, highly depends on the experience of the radiologist and due to the increasing number of images per study can be time-consuming.

Referred to as ``Radiomics''~\cite{Gillies:2016,Kumar:2012,Lambin:2012, Oikonomou:2018}, the ability to process such large amounts of data promises to decipher the encoded information within medical images; Develop predictive and prognostic models to design personalized diagnosis; Allow comprehensive study of tumor phenotype~\cite{Aerts:2014}, and; Assess tissue heterogeneity for diagnosis of different type of cancers. More specifically, Radiomics refers to the process of extracting and analyzing several semi-quantitative (e.g., attenuation, shape, size, and location) and/or quantitative features (e.g., wavelet decomposition, histogram, and gray-level intensity)  from medical images with the ultimate goal of obtaining predictive or prognostic models.

Although several challenges are in the way of bringing Radiomics into daily clinical practice, it is expected that Radiomics become a critical component for integration of image-driven information for personalized treatment in the near future. 

It is worth mentioning that computer aided diagnosis (CAD) is not a new concept, and researchers have developed automatic systems to investigate the link between imaging-based features and  biological characteristics in the past. However, this field is formalized as ``Radiomics'', since $2010$~\cite{Gillies:2010}, and it has a few key differences with the traditional CAD systems. First of all, CAD systems incorporate much fewer number of features (typically, within $8$ to $20$ features), whereas in Radiomics, hundreds to thousands of features are extracted. Second, the application of the CAD systems is, typically, limited to the diagnosis of the diseases, such as distinguishing between benign and malignant masses. Nevertheless, Radiomics is a much broader field, including both predictive and prognostic applications~\cite{Avanzo:2017}. The first comprehensive clinical application of Radiomics was performed by Aerts \textit{et al.}~\cite{Aerts:2014} with involvement of $1019$ lung cancer patients. More than $400$ different intensity, shape, texture, and wavelet features were extracted from Computed Tomography (CT)  images and used together with clinical information and gene expression data to develop Radiomics heat map, which shows the association between Radiomics and different clinical outcomes such as cancer stage.  This clinical study has illustrated/validated  effectiveness of Radiomics for tumor related predictions and showed that Radiomics has the capability to identify lung and head-and-neck cancers from a single-time point CT scan. Consequently, there has been a recent surge of interest~\cite{Zhang:2017,Griethuysen:2017,Sun:2017,Echaniz:2017,Sun:2016} on this multidisciplinary research area as Radiomics has the  potential to provide significant assistance for assessing the risk of recurrence of cancer~\cite{HLi:2016}; Evaluating the risk of radiation-induced side-effects on non-cancer tissues~\cite{Cunliffe:2015}, and; Predicting the risk for cancer development in healthy subjects~\cite{Cunliffe:2015}. In a very recent article by Vallieres \textit{et al.}~\cite{Vallieres:2018}, it is shown that the Radiomics features extracted in Reference~\cite{Aerts:2014} have a noticeable dependency on the tumor volume, which is a strong prognostic factor, and revised calculations are proposed that are less correlated to the tumor volume. In other words, more powerful Radiomics features and procedures are being introduced, illustrating the ongoing research potentials of Radiomics.

The key underlying hypothesis in the Radiomics is that the constructed descriptive models (based on medical imaging data, sometimes complemented by biological and/or medical data) are capable of providing relevant and beneficial predictive, prognostic, and/or diagnostic information. In this regard, one can identify two main categories of Radiomics. Conventional pipeline based on Hand-Crafted Radiomic features ($\HCR$)  that consists of the following four main processing tasks: (i) Image acquisition/reconstruction; (ii) Image segmentation; (iii) Feature extraction and quantification, and; (iv) Statistical analysis and model building. On the other hand,  the Deep Learning-based Radiomics ($\DLR$) pipeline has recently emerged which differs from the former category since deep networks do not necessarily need the segmented Region Of Interest (ROI), and their feature extraction and analysis parts are partially or fully coupled. We will elaborate on these properties in section~\ref{sec:DL_radiomics}.

\begin{figure}[t!]
\centering
\includegraphics[width=0.5\textwidth]{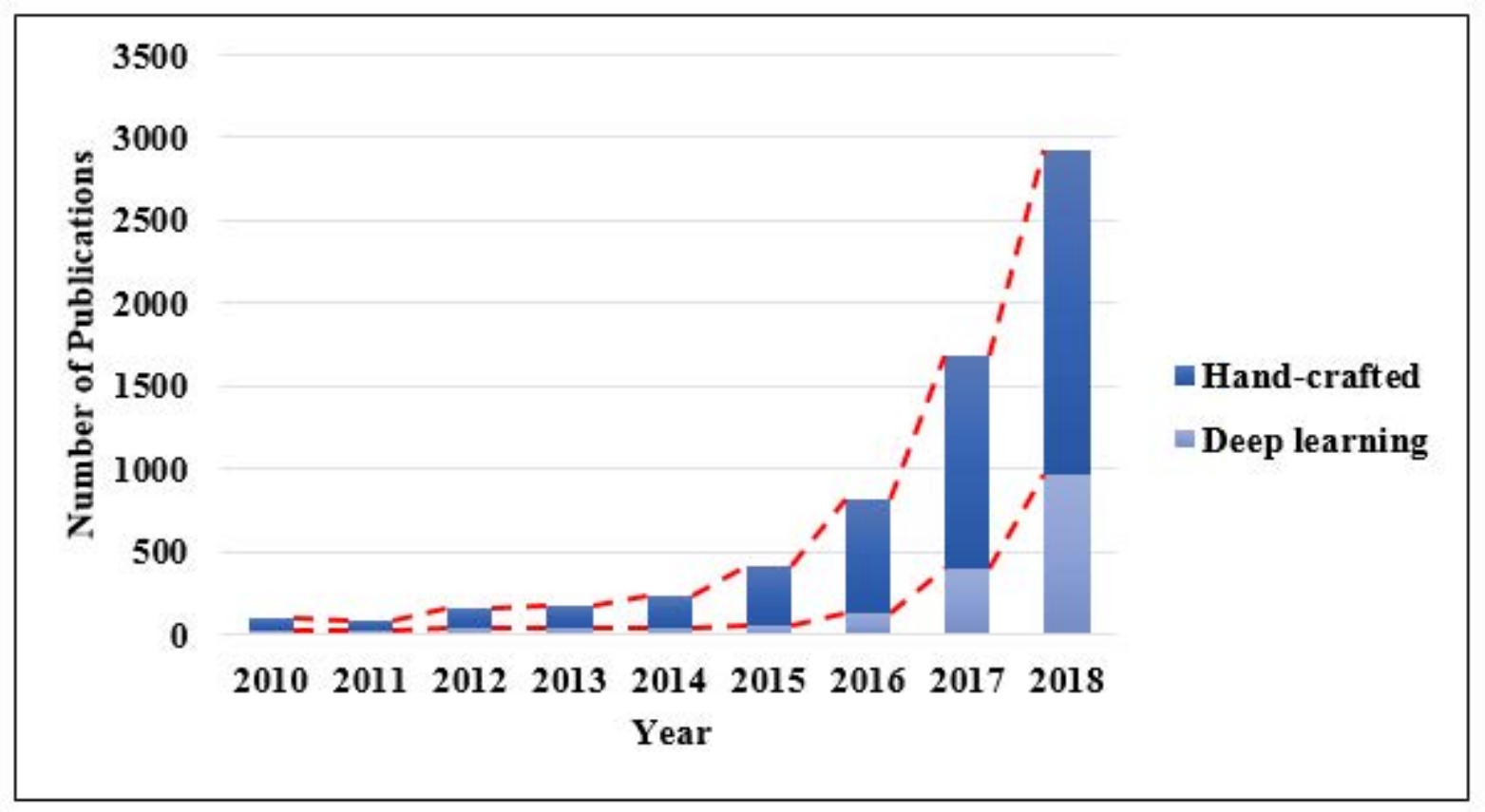}
 \centering
\caption{\small Increasing interest in Radiomics based on data from Google Scholar (``Radiomics'' is used as the keyword). It is observed that there is an increasing interest in both types of the Radiomics.} \label{fig:trend}
\vspace{-.2in}
\end{figure}
More clinical studies are being approved and conducted to further investigate and advance the unparalleled opportunities the Radiomics posed to offer for clinical applications.
Information Post~\ref{infoPost1} provides an overview of different screening technologies used within the Radiomics pipeline along with supporting data sources and available datasets to develop Radiomics-based predictive/prognostic models. While Radiomics consists of a wide range of (partially interconnected) research areas with each individual branch possibly worth a complete exploration, the purpose of this article is to provide an inclusive introduction to Radiomics for the signal processing community, as such, we will focus on the progression of signal processing algorithms within the context of Radiomics. In brief, we aim to present an overview of the current state, opportunities, and challenges of Radiomics from  signal processing perspective to facilitate further advancement and innovation in this critical multidisciplinary research area. As such, the article will cover the following four main areas:
\begin{enumerate}
\item[(i)] Hand-crafted Radiomics, where we introduce and investigate different feature extraction, feature reduction, and classification approaches used within the context of Radiomics. Most of the techniques utilized in any of the aforementioned steps lie within the broad area of ``Machine Learning,'' where the goal is to improve the performance of different computational models using past experiences (data)~\cite{PGrossmann:2015}. In other words, the underlying models are capable of learning from past data, leading to the automatic process of prediction and diagnosis. Furthermore, since hundreds of Radiomics features are extracted, an appropriate feature selection/extraction strategy should be adopted to reduce the ``curse of dimensionality'' and overfitting of the prediction models. Most of these strategies, themselves, lie within the field of ``Machine Learning'', as they are aimed to learn the best set of features, based on the available data.
\item[(ii)] Deep learning-based Radiomics, where we provide an overview of different deep architectures used in Radiomics along with interpretability requirements.
\item[(iii)] Hybrid solutions developed to simultaneously benefit from the advantages of each of the above two mentioned  categories.
\item[(iv)] Challenges, Open Problems, and Opportunities, where we focus on the limitations of processing techniques  unique in nature to the Radiomics, and introduce open problems and potential opportunities for signal processing researchers.
\end{enumerate}
Fig.~\ref{fig:trend} shows the increasing interest in Radiomics within the research community. Although there have been few recent articles~\cite{Tian:2018, Gillies:2016} reviewing and introducing Radiomics,  to the best of our knowledge, most of them are from outside the signal processing (SP) community. References within the SP society such as the work by J. Edwards~\cite{Edwards:2017} have investigated recent advancements in medical imaging devices and technologies without reviewing the role of Radiomics in medical applications.
Other existing papers outside SP community (e.g.,~\cite{Gillies:2016}) have failed to clearly describe the underlying signal processing technologies and have narrowed down their scope only to hand-crafted Radiomics and its diagnosis capability. While Reference~\cite{Tian:2018} has briefly touched upon the deep learning pipeline as an emerging technology that can extract Radiomics features, it has not studied applicability of different deep architectures~\cite{Ravi:2017} and left the interpretability topic untouched. Furthermore, Reference~\cite{Constanzo:2017} has mostly focused on the hand-crafted Radiomics, while  deep learning-based Radiomics is explained briefly without addressing different architectures, interpretability, and hybrid models. Although both types of Radiomics are covered in Reference~\cite{CParmar:2018}, combination of hand-crafted and deep learning-based features are not considered. Besides, challenges associated with Radiomics and the relation between Radiomics and gene-expression (Radiogenomics) are also not discussed thoroughly. Finally, the scope of Reference~\cite{Litjens:2017} is limited to deep learning-based Radiomics, without addressing hand-crafted features, their stability, hybrid Radiomics, and Radiogenomics. All these call for an urgent and timely quest to introduce Radiomics to our community especially since SP is one of the main building blocks of the Radiomics.

The reminder of this article is organized as follows: first in Section~\ref{sec:App}, we will discuss several applications of Radiomics in cancer-related fields, followed by Hand-Crafted solutions in Section~\ref{sec:HC_radiomics}. The Deep learning-based Radiomics is presented in Section~\ref{sec:DL_radiomics}, where several aspects of $\DLR$ is investigated. In Section~\ref{sec:HybSol}, we explain different hybrid solutions to Radiomics, which aim to take advantage of both $\DLR$ and $\HCR$. Finally in Section~\ref{sec:COO} various challenges and opportunities of Radiomics, especially for SP community, are discussed. We conclude our work is Section~\ref{sec:Conc}.

\begin{strip}
\begin{tcolorbox}[colback=green!5,colframe=green!40!black,title=Information Post I: Radiomics Supporting Resources,label=infoPost1]
\begin{multicols}{2}
\small
Several potential medical resources provide information to the Radiomics pipeline, some of which are directly used to extract Radiomics features and some serve the decision making process, as complementary information sets. Below we review the most important data resources for Radiomics.

\vspace{.1in}
\noindent
\textbf{Screening Technologies}: The Radiomics features can be extracted from several imaging modalities, among which the following are the most commonly used modalities:
\begin{itemize}
\item \textit{\textbf{Computed Tomography (CT) Scans}}: The CT is the modality of choice for the diagnosis of many diseases in different parts of the body, and by providing high resolution images~\cite{Lambin:2012} paves the path for extracting comparable Radiomics features. Nonetheless, the CT imaging performance depends on different components of the utilized protocol including the following three main properties: (i) Slice thickness, which is the distance in millimeter (\textit{mm}) between two consecutive slices; (ii)  The capability for projecting the density variations into image intensities, and; (iii)~Reconstruction algorithm, which aims at converting tomographic measurements to cross-sectional images. Although CT protocols for specific clinical indications are usually similar across different institutions, Radiomics features can even differ between different scanners with the same settings~\cite{Berenguer:2018}. Therefore, there is still a considerable need to ensure consistency of Radiomics feature extraction amongst different scanners and imaging protocols~\cite{Kumar:2012}. CT images are typically divided into two categories~\cite{Thawani:2017}: screening and diagnostic. While screening CT uses low dose images, diagnostic CT utilizes high dose and is of higher quality and contrast.
\item \textit{\textbf{Positron Emission Tomography (PET) Scans}}: The PET is a nuclear imaging modality that evaluates body function and metabolism \cite{Lambin:2012}, and since its performance depends on not only the scanner properties, but also the doze calibration, similar to the case with the CT scans, standardizing the PET protocols across different institutions is challenging. Furthermore, glucose level at the time of scanning can also affect the properties of PET images~\cite{Kumar:2012}.
\columnbreak
\item \textit{\textbf{Magnetic Resonance Imaging (MRI)}}: Unlike CT, properties of MRI images are not directly associated with tissue density and specific methods are required to obtain the so-called signal intensity. Besides, several imager and vendor-dependant factors such as gradient and coil systems~\cite{Sasaki:2008}, pulse sequence design, slice thickness, and other parameters such as artifacts and magnetic field strength affect the properties of the MRI images~\cite{Kumar:2012}, which should be consistent across different institutions.
\end{itemize}
\begin{center}
\includegraphics[width=8cm]{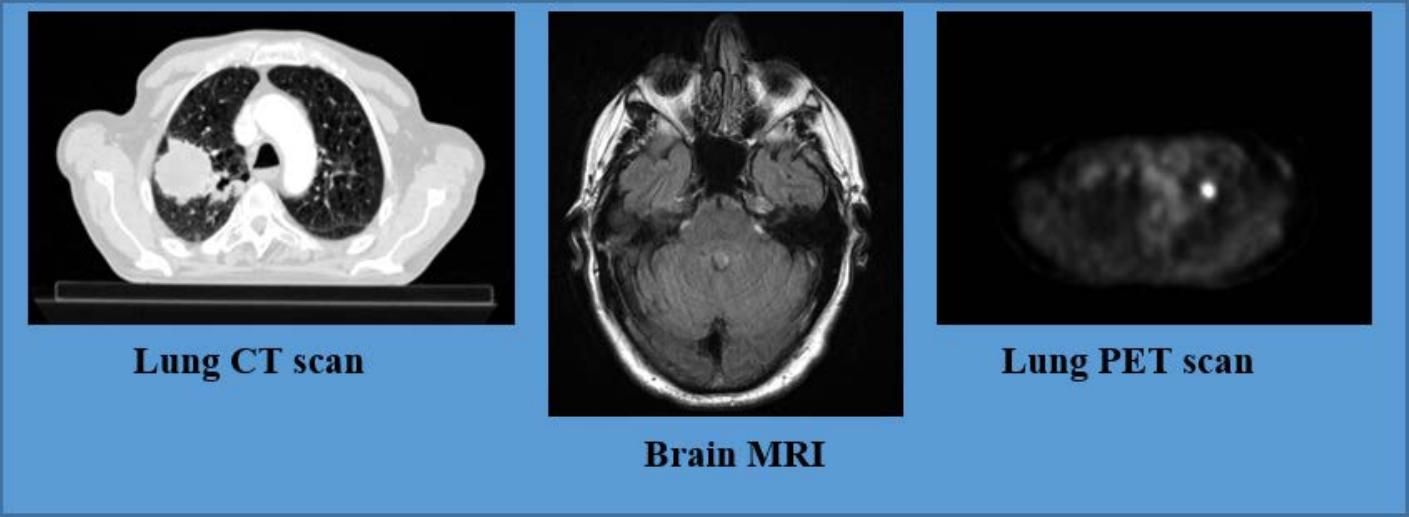}
\end{center}
\vspace{.1in}
\noindent
\textbf{Complimentary Data Sources}: In addition to imaging resources, the following clinical data sources are typically combined with Radiomics features:

\begin{itemize}
\item \textit{\textbf{Gene expression}}: The process of converting DNA to functional product to have a global insight of cellular function.
\item \textit{\textbf{Clinical characteristics}}: Patient's characteristics such age, gender, and past medical and family history~\cite{Kumar:2012}.
\item \textit{\textbf{Blood Bio-markers}}: Measurable characteristics from the patient's blood such as glucose level, cholesterol level and blood pressure.
\item \textbf{\textit{Prognostic Markers}}: Markers to evaluate the progress of the disease, response to treatment or survival, such as size, tumor stage, tumor recurrence, and metastasis.
\end{itemize}

\vspace{.1in}
\noindent
\textbf{Radiomics Image Databases}: Large amount of data is typically required to get reliable results about tissue heterogeneity based on Radiomics~\cite{Kumar:2012}. Table~\ref{tab:data} below introduces a few publicly available imaging data source that can be used to develop and test new Radiomics approaches:
\end{multicols}

\begin{table}[H]

\centering
\caption{Popular data sets for performing Radiomics.}
\begin{tabular}{ccccccc}

 \textbf{Data Set} & \textbf{Reference} & \textbf{Year} & \textbf{Imaging Modality} & \textbf{Type of the Tumor} & \textbf{Number of Patients} & \textbf{Annotation Type} \\
\midrule
LIDC-IDRI & \cite{lidc:2015} & 2015 & CT & Lung Tumor & $1010$ & Multiple experts\\
NSCLC-Radiomics & \cite{Aerts:2014} &2014 & CT& Lung Tumor &$422$& One expert\\
NSCLC-Radiomics-Genomics & \cite{Aerts:2014} & 2014 & CT & Lung tumor &$89$& -\\
LGG-1p19qDeletion& \cite{Akkus:2017} & 2017 & MRI & Brain Tumor &$159$ & One expert\\
Head-Neck-PET-CT& \cite{Vallieres:2017} & 2017 & PET, CT & Head-and-Neck Cancer &$298$& -\\
BRATS2015&\cite{Menze:2015}&2015&MRI&Brain Tumor&$65$&Multiple experts
\end{tabular}
\label{tab:data}

\end{table}
\end{tcolorbox}
\end{strip}

\section{Applications} \label{sec:App}
\begin{table*}[ht!]
\centering
\small
\caption{Applications of Radiomics.}\label{tab:app1}
\begin{tabularx}{\textwidth}{XsYXYsY}
\arrayrulecolor{LightCyan}
\rowcolor{LightCyan}
\textbf{} \!\!\!&\!\!\! \textbf{Imaging} \!\!\!&\!\!\! \textbf{Information} \!\!\!&\!\!\! \textbf{~~~Application} \!\!\!&\!\!\! \textbf{Radiomics}\!\!\!&\!\!\! \textbf{Number of}\!\!\!&\!\!\! \textbf{Type of}\\
\rowcolor{LightCyan}
\textbf{~~~~~~Reference} \!\!\!&\!\!\! \textbf{Modality} \!\!\!&\!\!\! \textbf{Sources} \!\!\!&\!\!\! \textbf{~~~~~Domain} \!\!\!&\!\!\! \textbf{Method}\!\!\!&\!\!\! \textbf{Patients}\!\!\!&\!\!\! \textbf{Annotation} \\
\arrayrulecolor{LightCyan}
Zhang {\em et al.}\cite{Zhang:2017}  \!\!\!&\!\!\! CT\!\!\!&\!\!\! \centering{-} \!\!\!&\!\!\! Prediction of lung cancer recurrence and death.\!\!\!&\!\!\! Hand-crafted Radiomics ($\HCR$)&$112$&One expert\\
\\
Aerts {\em et al.}\cite{Aerts:2014}\!\!\!&\!\!\! CT \!\!\!&\!\!\!Gene-expression and Clinical data.\!\!\!&\!\!\! Lung, and head \& neck cancer survival prediction. \!\!\!&\!\!\! $\HCR$&$1019$&Multiple experts\\
\\
Griethuysen {\em et al.}\cite{Griethuysen:2017}\!\!\!&\!\!\!CT\!\!\!&\!\!\!\centering{-}\!\!\!&\!\!\! Lung cancer benign and malignant classification.\!\!\!&\!\!\! $\HCR$&$302$&Multiple experts\\
\\
Oikonomou {\em et al.}\cite{Oikonomou:2018} \!\!\!&\!\!\! CT, PET \!\!\!&\!\!\!Standardized uptake value (ratio of image and body radioactivity concentration).\!\!\!&\!\!\!Lung cancer survival prediction. \!\!\!&\!\!\! $\HCR$&$150$&One expert\\
\\
Kumar {\em et al.}\cite{Kumar:2015}\!\!\!&\!\!\!CT\!\!\!&\!\!\!\centering{-}\!\!\!&\!\!\!Lung cancer benign and malignant classification.\!\!\!&\!\!\! Deep Learning-based Radiomics ($\DLR$)&$1010$&Multiple experts\\
\\
Kumar {\em et al.}\cite{Chung:2017}\!\!\!&\!\!\!CT\!\!\!&\!\!\!\centering{-}\!\!\!&\!\!\!Lung cancer benign and malignant classification.\!\!\!&\!\!\! $\DLR$&$97$&Multiple experts\\
\\
Huynh {\em et al.}\cite{Li:2016}\!\!\!&\!\!\!Mammogram\!\!\!&\!\!\!\centering{-}\!\!\!&\!\!\!Classification of breast cancer: benign or malignant.\!\!\!&\!\!\! $\HCR$, $\DLR$, Combination of $\HCR$ and $\DLR$&$219$&Semi-automatic\\
\\
Li {\em et al.}\cite{Li:2017}\!\!\!&\!\!\!MRI\!\!\!&\!\!\!\centering{-}\!\!\!&\!\!\!IDH1 enzyme mutation prediction.\!\!\!&\!\!\! $\DLR$&$151$&Automatic\\
\\
Sun {\em et al.}\cite{Sun:2017}\!\!\!&\!\!\!CT\!\!\!&\!\!\!\centering{-}\!\!\!&\!\!\!Lung cancer benign and malignant classification.\!\!\!&\!\!\!$\HCR$, $\DLR$&$1018$&Multiple experts\\
\\
Jamaludin {\em et al.}\cite{Jamaludin:2016}\!\!\!&\!\!\!MRI\!\!\!&\!\!\!\centering{-}\!\!\!&\!\!\!Disc abnormality classification.\!\!\!&\!\!\!$\DLR$&$2009$&\\
\\
Liu {\em et al.}\cite{Liu:2017}\!\!\!&\!\!\!MRI\!\!\!&\!\!\!\centering{-}\!\!\!&\!\!\!Prostate cancer diagnosis.\!\!\!&\!\!\! $\HCR$, $\DLR$&$341$&Not performed\\
\\
Oakden-Rayner {\em et al.}\cite{Oakden-Rayner:2017}\!\!\!&\!\!\!CT\!\!\!&\!\!\!\centering{-}\!\!\!&\!\!\!Longevity prediction.\!\!\!&\!\!\!$\HCR$, $\DLR$&$48$&Semi-automatic\\
\\
Paul {\em et al.}\cite{Hawkins:2016} \!\!\!&\!\!\! CT \!\!\!&\!\!\!\centering{-}\!\!\!&\!\!\!Lung cancer short/long-term survival prediction.\!\!\!&\!\!\! Combination of $\HCR$ and $\DLR$&$81$&Semi-automatic\\
\\
Fu {\em et al.}\cite{Fu:2017}\!\!\!&\!\!\!CT\!\!\!&\!\!\!\centering{-}\!\!\!&\!\!\!Lung tumor detection.\!\!\!&\!\!\!Combination of $\HCR$ and $\DLR$&$1010$&Not performed\\
\\
Bickelhaupt {\em et al.}\cite{Bickelhaupt:2017}\!\!\!&\!\!\!Mammogram\!\!\!&\!\!\!\centering{-}\!\!\!&\!\!\!Classification of breast cancer: benign or malignant.\!\!\!&\!\!\!$\HCR$&$50$&One expert\\
\\
\arrayrulecolor{LightCyan}\midrule
\end{tabularx}
\end{table*}

\begin{table*}[ht!]
\centering
\small
\caption{Applications of Radiomics (Continued).}\label{tab:app2}
\begin{tabularx}{\textwidth}{XYYXYYY}
\arrayrulecolor{LightCyan}
\rowcolor{LightCyan}
\textbf{} \!\!\!&\!\!\! \textbf{Imaging} \!\!\!&\!\!\! \textbf{Information} \!\!\!&\!\!\! \textbf{~~~Application} \!\!\!&\!\!\! \textbf{Radiomics}\!\!\!&\!\!\! \textbf{Number of}\!\!\!&\!\!\! \textbf{Type of}\\
\rowcolor{LightCyan}
\textbf{~~~~~~Reference} \!\!\!&\!\!\! \textbf{Modality} \!\!\!&\!\!\! \textbf{Sources} \!\!\!&\!\!\! \textbf{~~~~~Domain} \!\!\!&\!\!\! \textbf{Method}\!\!\!&\!\!\! \textbf{Patients}\!\!\!&\!\!\! \textbf{Annotation} \\
\arrayrulecolor{LightCyan}
Lao {\em et al.}\cite{Lao:2017}\!\!\!&\!\!\! MRI\!\!\!&\!\!\! Clinical risk factors.\!\!\!&\!\!\!Brain cancer survival prediction.\!\!\!&\!\!\! Combination of $\HCR$ and $\DLR$&$112$&Multiple experts\\
\\
Antropova {\em et al.}\cite{Antropova:2017}\!\!\!&\!\!\!Mammogram, Ultrasound, MRI\!\!\!&\!\!\!\centering{-}\!\!\!&\!\!\!Breast cancer benign and malignant classification.\!\!\!&\!\!\!Combination of $\HCR$ and $\DLR$&$2060$&Semi-automatic\\
\\
Wang {\em et al.}\cite{Zhou:2017}\!\!\!&\!\!\!CT, PET\!\!\!&\!\!\!Standardized uptake value.\!\!\!&\!\!\!Lung cancer benign and malignant classification.\!\!\!&\!\!\!$\HCR$, $\DLR$&$168$&One expert\\
\\
Shen {\em et al.}\cite{WShen:2017}\!\!\!&\!\!\!CT\!\!\!&\!\!\!\centering{-}\!\!\!&\!\!\!Prediction of lung tumor malignancy likelihood.\!\!\!&\!\!\!$\DLR$&$1010$&Not performed\\
\\
Emaminejad {\em et al.}\cite{Emaminejad:2016}\!\!\!&\!\!\!CT\!\!\!&\!\!\!Genomics bio-markers.\!\!\!&\!\!\!Lung cancer recurrence risk prediction\!\!\!&\!\!\!$\HCR$&$79$&Semi-automatic\\
\\
Sun {\em et al.}\cite{Sun:2016}\!\!\!&\!\!\!CT\!\!\!&\!\!\! \centering{-}\!\!\!&\!\!\!Lung cancer benign and malignant classification.\!\!\!&\!\!\!$\HCR$, $\DLR$&$1018$&One expert\\
\\
Kim {\em et al.}\cite{Kim:2016}\!\!\!&\!\!\!CT\!\!\!&\!\!\! \centering{-}\!\!\!&\!\!\!Lung cancer benign and malignant classification.\!\!\!&\!\!\!Combination of $\HCR$ and $\DLR$&$20$&One expert\\
\\
Shen {\em et al.}\cite{WShen:2016}\!\!\!&\!\!\!CT\!\!\!&\!\!\! \centering{-}\!\!\!&\!\!\!Lung cancer malignancy probability estimation.\!\!\!&\!\!\!$\DLR$&$1010$&Not performed\\
\\
Ciompi {\em et al.}\cite{Ciompi:2017}\!\!\!&\!\!\!CT\!\!\!&\!\!\! \centering{-}\!\!\!&\!\!\!Lung tumor classification as solid and non-solid.\!\!\!&\!\!\!$\DLR$&$1411$&Not performed\\
\\
Afshar {\em et al.}\cite{Parnian:ICIP18-p1}\!\!\!&\!\!\!MRI\!\!\!&\!\!\! \centering{-}\!\!\!&\!\!\!Brain tumor type classification.\!\!\!&\!\!\!$\DLR$&$233$&One expert\\
\\
Cha {\em et al.}\cite{Cha:2017}\!\!\!&\!\!\!CT\!\!\!&\!\!\!\centering{-}\!\!\!&\!\!\!Bladder cancer treatment response prediction.\!\!\!&\!\!\!$\DLR$&$123$&Automatic\\
\\
Yu {\em et al.}\cite{Yu:2017}\!\!\!&\!\!\!CT\!\!\!&\!\!\!\centering{-}\!\!\!&\!\!\!Kidney tumor type classification.\!\!\!&\!\!\!$\HCR$&$119$&Multiple experts\\
\\
Zhou {\em et al.}\cite{YZhou:2017}\!\!\!&\!\!\!CT\!\!\!&\!\!\!\centering{Gene expression.}\!\!\!&\!\!\!Liver cancer recurrence prediction.\!\!\!&\!\!\!$\HCR$&$215$&One expert\\
\\
\arrayrulecolor{LightCyan}\midrule
\end{tabularx}
\end{table*}
It is undeniable that automatic diagnosis systems are still in their infancy, and there is a long way before they can be reliably used in clinical applications. Having said that, several recent studies have investigated automatic diagnosis systems and compared them against human experts. For instance, Esteva \textit{et al.}~\cite{Esteva:2017} have developed a deep CNN  for skin cancer classification, using a dataset of $129,450$ clinical images. The performance of this system is tested against $21$ board-certificated experts, and results show that the performance of the system is on a par with human experts. This study suggests that the automatic diagnosis systems have the potential to achieve a human-level performance and can be utilized as one of the two experts. Similar to other emerging technologies, automatic diagnosis systems have different advantages and disadvantages. For instance, these systems have the potential to improve the quality of clinical care and decrease the number of medical errors. These systems, however, are associated with a major concern, i.e., the risk of violating patients' privacy, calling for strict regulations to ensure the privacy of the clinical information~\cite{Menachemi:2011}.

In recent years, Radiomics has been applied to many health-care applications, including oncology, cardiology, and neurology. In cardiology, for instance, Radiomics is used in different investigations, such as identifying the coronary plaques~\cite{Kolossvary:2017}. In neurology, it is widely applicable for detecting Alzheimer's disease~\cite{Suk:2014} and Parkinson's disease~\cite{Rahmim:2017}. However, among all the applications of the Radiomics, cancer-related topics have been the focus of interest. Below we briefly introduce and define different cancer-related applications in which Radiomics has been shown to be successful.
\begin{enumerate}
\item \textit{\textbf{Cancer diagnosis}}, which refers to confirming the presence or absence of the cancer, is one of the most critical and sensitive decisions that has to be made as early as possible. However, most of the times cancers are diagnosed in late stages reducing the chance of receiving effective treatment, as there are typically few clinical symptoms in the early stages of cancer. Nevertheless, Radiomics has the potential to improve the accuracy of cancer early diagnosis.
\item \textit{\textbf{Tumor detection}} refers to the identification of those lesions that are malignant, which is very important in order to guide targeted local treatment. For instance, Radiotherapy, the process of killing cancerous cells using ionizing radiation, can have much more efficient results if it is focused on the more ``aggressive" areas of the tumor (usually the more heterogeneous areas are the more aggressive ones. Drug delivery, i.e., having an exact plan to deliver the drug to the target area, is another problem that requires precise information about the abnormality location.
\item \textit{\textbf{Tumor classification and attribute scoring}}: Tumor classification refers to determining the type of the tumor. Typically, cancer is classified into the following main classes: (i) benign; (ii) primary malignant, and; (iii) metastatic malignant. Besides, tumors are associated with different attributes such as their border and sphericity. Analyzing these attributes contribute to a better understanding of the tumor's shape and behavior.
\item \textit{\textbf{Survival prediction}}: The knowledge of the expected survival of a specific disease with or without a specific treatment is critical both for treating physicians and the patients. Physicians need to choose the best treatment plan for their patients and patients need to know their predicted survival time in order to make their own choices for the quality of their life. Radiomics can add significant information about patient's survival based on image properties and heterogeneity of the tumor and this has attracted a lot of attention recently.
\item \textit{\textbf{Malignancy prediction}}: Tumors can be either malignant or benign based on several factors such as their ability to spread to other tissues. Benign tumors usually do not spread to other organs but may need surgical resection because occasionally they may grow in size. Pre-invasive lesions may be indolent for years, however, they may transform to aggressive malignant tumors and therefore need to be monitored closely or even be treated with lower dose of anti-cancer regimens. Malignant tumors are life threatening and may spread to distant organs, requiring more complicated treatments such as Chemotherapy. Prediction of tumor malignancy likelihood with noninvasive methods such as Radiomics is, therefore, of paramount importance.
\item \textit{\textbf{Recurrence prediction}}: Even the treated cancers have the potential to grow and reappear, which is referred to as ``cancer recurrence''. As the cancerous region is supposed to be removed or treated, there are not strong landmarks or evidences helping with predicting the recurrence. However, recently Radiomics is being employed to assist with such issue and has shown promising initial results.
\item \textit{\textbf{Cancer staging}}: Cancers may be diagnosed in different stages, e.g., they may be in an early stage meaning that they are remaining in the tissue they have first appeared in, or they can be in an advanced stage, meaning that they are spread in other tissues. Knowing the stage of the tumor has significant impact on the choice of required treatment.
\end{enumerate}
Based on the above categories, Tables~\ref{tab:app1} and~\ref{tab:app2} summarize different application domains of Radiomics introduced in various articles, along with their associated Radiomics method ($\HCR$, $\DLR$, or the combination of both). These tables also provide information on any complementary data source that has been utilized in combination with Radiomics.

\section{State-of-the-Art in Hand-Crafted Radiomics} \label{sec:HC_radiomics}
In clinical oncology, tissue biopsy, which refers to the removal of a small focal part of the cancerous tissue (tumor), is considered as the state-of-the-art approach for diagnosing cancer. Although tissue biopsy has several diagnostic properties and is widely used for detecting and investigating cancerous cells, its reliability is limited by the fact that tumors are spatially and temporally heterogeneous, and as a result, biopsy cannot capture all the available  information that is necessary for an inclusive decision. Besides, most of the biopsies are invasive which restricts the number of times this procedure can be performed, or sometimes biopsy is not an option due to the high risk for complication that pertains to specific patients.

Although biopsy remains the gold standard for cancer diagnosis, it can be combined with Radiomics, which is a non-invasive technique and can capture intra-tumoral heterogeneity. The resulting imaging-guided biopsy is a much less interventional procedure associated with fewer complications compared to surgical biopsy. In other words Radiomics can be used to facilitate biopsy by detecting more suspicious locations~\cite{Gillies:2016}. Furthermore, Radiomics can provide complementary information for diagnosis, and in case of ``negative result of a biopsy'' Radiomics prediction models may also provide additional information to the clinicians on whether re-biopsy is needed~\cite{YLiu:2016}.

In this section, we focus on the state-of-the-art research on hand-crafted Radiomics ($\HCR$).  Studies on hand-crafted Radiomics features ~\cite{Kumar:2012,Lambin:2012,Oikonomou:2018}, typically, consist of the following key steps:
\begin{enumerate}
\item[1.] \textit{\textbf{Pre-processing}}, introduced to reduce noise and artifacts from the original data and typically includes image smoothing and image enhancement techniques.
\item[2.] \textit{\textbf{Segmentation}}, which is a critical step within the $\HCR$ workflow, as typically $\HCR$ features  are extracted from segmented sections and many tissues do not have distinct boundaries~\cite{Gillies:2016}. Although manual delineation of the gross tumor is the conventional (standard) clinical approach, it is time consuming and extensively sensitive to inter-observer variability~\cite{Kumar:2012}, resulting in a quest to develop advanced (semi) automated segmentation solutions of high accuracy that can also generate reproducible boundaries.

Automatic and semi-automatic segmentation techniques can be either conventional, meaning that pre-defined features are used to classify image pixels/voxels as tumorous or non-tumorous, or deep learning-based, referring to the use of a deep network to segment the image. Conventional techniques can, themselves, lie within three categories of intensity-based~\cite{Dehmeshki:2008}, model-based, and machine learning methods. In the former category, intensity is used as the main distinguishing feature of the pixels, while in the model-based approaches, the aim is to improve an initial contour, by optimizing an energy function. In machine learning methods, however, a set of features, including intensity and gradient, are extracted from the pixels. These features are then used as the inputs to a machine learning model, such as a Support Vector Machine (SVM), to classify the pixels. Nevertheless, conventional techniques are subject to several shortcomings. For instance, the intensity of the tumor can, sometimes, be similar to other tissues, and therefore, intensity can not be a good discriminator. Furthermore, the formulation of an energy function, in a model-based segmentation, may involve large number of parameters~\cite{Farag:2013}, which makes optimization of the energy function difficult and time-consuming. Deep learning methods, on the other hand, are capable of learning the features that can best distinguish tumorous and non-tumorous pixels, and can be trained in an end-to-end manner. Deep learning approaches, such as different variations of the U-Net~\cite{Ronneberger:2015}, ``LungNet'' architecture~\cite{Anthimopoulos:2018}, DenseNet~\cite{Huang:2017}, and hybrid dilated convolutions (HDC)~\cite{Wang:2018} are currently used more often for medical image segmentation.

The most important metric for evaluating a segmentation method is to calculate its accuracy according to a ground truth. However, since ground truth is not always available for medical images, reproducibility metrics are often used to assess the performance of the segmentation algorithm~\cite{Kumar:2012,Gillies:2016}.  For instance, reference~\cite{Kumar:2012} has used a similarity metric based on the overlap of generated segments resulting in a better average for automatic methods compared with manual delineation.
\item[3.] \textit{\textbf{Feature extraction}}, which is the main step in Radiomics workflow and will be discussed in details in Sub-section~\ref{RFE}.
\item[4.] \textit{\textbf{Feature reduction}}, is another critical step in Radiomics as although a large number of quantitative features can be extracted from the available big image datasets, most of the features are highly correlated, irrelevant to the task at hand, and/or contribute to over-fitting of the model. To address these issues, Radiomics feature reduction techniques are discussed in Sub-section~\ref{sec:RFR}.
\item[5.] \textit{\textbf{Statistical analysis}}, which refers to utilizing the extracted Radiomics features in a specific application as outlined in Section~\ref{sec:App}. We will further elaborate on such Radiomics-based statistical analysis in Sub-section~\ref{sec:RSA}.
\end{enumerate}
In the reminder of this section, we focus on Steps 3-5 in Sub-sections~\ref{RFE}-\ref{sec:RSA}, respectively, starting by reviewing  the key feature extraction methodologies recently used in $\HCR$, followed by a review of the main feature reduction techniques and Radiomics-based statistical analytics.

\subsection{Radiomics Feature Extraction} \label{RFE}
\begin{table*}[t!]
\small
\begin{center}
\caption{Different categories of $\HCR$ features commonly used within the context of Radiomics.}
\begin{tabularx}{\textwidth}{XXX}
\arrayrulecolor{LightCyan}\hline
\rowcolor{LightCyan}
\textbf{~~~~~~~~~~~~~~~~~~~Category} & \textbf{~~~~~~~~~~~~~~~~~~~Description} & \textbf{~~~~~~~~~~~~~~~~~~~Sub-category} \\

First Order Radiomics  & Concerned with the distribution of pixel intensities and use of elementary metrics to compute geometrical features. & \\
\\
~~$\bullet$ Shape Features & Quantify the geometric shape of region or volume of interest ~\cite{Kumar:2012}&
Size of the Region of Interest (ROI); Sphericity; Compactness; Total volume; Surface area, Diameter, flatness and; Surface-to-volume ratio~\cite{Kumar:2012,Thawani:2017}.\\\\
~~$\bullet$  Intensity Features & Derived from a single histogram generated from the 2D region or the whole 3D volume~\cite{Kumar:2012}. & Intensity Mean; Intensity Standard Deviation; Intensity Median; Minimum of Intensity; Maximum of Intensity; Mean of Positive Intensities; Uniformity; Kurtosis; Skewness; Entropy; Normalized Entropy;  Difference of Entropy; Sum of Entropy, and; Range~\cite{Kumar:2012,Thawani:2017}. \\
\\
Second Order Radiomics (Texture Features) & Concerned with texture features and relations between pixels to model intra-tumor heterogeneity.  Texture features are generated from different descriptive matrices~\cite{Kumar:2012}. & \\
\\
~~$\bullet$ Gray Level Co-occurrence~(GLCM) & GLCM~\cite{Thawani:2017}  is a matrix that presents the number of times that two intensity levels have occurred in two pixels with specific distance.& Contrast; Energy; Correlation; Homogeneity;Variance; Inverse Difference Moment; Sum of Average; Sum of Variance; Difference of Variance; Information Measure of Correlation; Autocorrelation; Dissimilarity; Cluster Shade; Cluster Prominence;  Cluster Tendency, and; Maximum Probability.\\
\\
~~$\bullet$ Gray Level Run-Length~(GLRLM) & GLRLM~\cite{Parekh:2016} is a matrix that presents the length of consecutive pixels having the same intensity.  &  Short run emphasis; Long run emphasis; Gray Level Non-Uniformity; Run length non-uniformity; Run percentage; Low gray level run emphasis, and; High gray level run emphasis~\cite{Kumar:2012}.\\
\\
~~$\bullet$  Neighborhood Gray Tone Difference Matrix (NGTDM)  & NGTDM~\cite{Thawani:2017} is concerned with the intensities of neighboring pixels instead of the pixel itself. & Coarseness; Contrast; Busyness; Complexity Texture Strength.\\
\\
~~$\bullet$  Grey-Level Zone Length Matrix (GLZLM)  & GLZLM~\cite{Griethuysen:2017} considers the size of homogeneous zones in every dimension. & Zone Percentage; Short-Zone Emphasis; Long-Zone Emphasis; Gray-Level Non-Uniformity for zone;  Zone Length Non-Uniformity.\\
\\
Higher Order Radiomics & Use of filters to extract patterns from images. & Wavelets; Fourier features~\cite{Thawani:2017};  Minkowski functionals; Fractal Analysis~\cite{Gillies:2016}, and; Laplacian of Gaussian (LoG)~\cite{Griethuysen:2017}.\\
\arrayrulecolor{LightCyan}\midrule
\end{tabularx}
\label{table:HCR-class}
\end{center}
\end{table*}

During the feature extraction step within Radiomics workflow, different types of features are extracted that can be generally classified into three main categories: (1) First order (intensity-based and shape-based features)~\cite{Zhang:2017}; (2) Second order (texture-based features)~\cite{Zhang:2017}, and; (3) Higher order features~\cite{Gillies:2016}. Table~\ref{table:HCR-class} provides a summary of different potential features. It is worth mentioning that $\HCR$ features are not limited to this list and can exceed hundreds of features (e.g., in Reference~\cite{Aerts:2014} 400 $\HCR$ features are initially extracted before going through a feature reduction process). Below, we further investigate the most commonly used categories of hand-crafted features:

\vspace{.1in}
\noindent
\textbf{1. Intensity-based Features}:
Intensity-based methods convert the multi-dimensional ROI into a single histogram (describing the distribution of pixel intensities), from which simple and basic features (e.g., energy, entropy, kurtosis, and skewness) are derived. Intensity features allow us to investigate properties of the histogram of tumor intensities such as sharpness, dispersion, and asymmetry. These features are, however, the most sensitive ones to image acquisition parameters such as slice thickness~\cite{Thawani:2017} (discussed in the Information Post~\ref{infoPost1}). Therefore, designing intensity-based features need special care and pre-processing. Among all intensity features, entropy and uniformity are the most commonly used ones in Radiomics~\cite{Parekh:2016}. Generally speaking, entropy measures the degree of randomness within the pixel intensities, and takes its maximum value when all the intensities occur with equal probabilities (complete randomness). Uniformity, on the other hand, estimates the consistency of pixel intensities, and takes its maximum value when  all the pixels are of the same value.

Although intensity-based features are simple to calculate and have the potential to distinguish several tissues such as benign and malignant tumors~\cite{Parekh:2016}, they suffer from some drawbacks. First, the selected number of bins can highly influence such features, as too small or too large bins can not resemble the underlying distribution correctly, and as such these features are not always reliable representatives. Besides, optimizing the number of histogram bins can also be problematic, because it leads to different number of bins for different ROIs, and makes it difficult to compare the results of various studies.

\vspace{.1in}
\noindent
\textbf{2. Shape-based Features}:
Shape-based  features describe the geometry of the ROI and are useful in the sense that they have high distinguishing ability for problems such as tumor malignancy and treatment response prediction~\cite{Thawani:2017}. Although radiologists commonly use shape features (also referred to as ``Semantic Features'' or ``Morphological features''), the aim of Radiomics is to quantify them with computer assistance~\cite{Gillies:2016}. These features are extracted from either 2D or 3D structures of the tumor region to investigate different shape and size characteristics of the tumor.

Among different shape-based features, volume, surface, sphericity, compactness, diameter, and flatness are more commonly used in Radiomics.  For instance, sphericity measures the degree of roundness of the volume or region of interest and it is specially useful for the prediction of tumor malignancy, as benign tumors are most of the times more sphere compared to malignant ones. Compactness is itself defined based on sphericity and as such, these two need not to be calculated simultaneously, and one of them will be probably excluded by the feature selection methods, which are targeting feature redundancy.

\vspace{.1in}
\noindent
\textbf{3. Texture-based Features}:
Shape-based  and intensity-based features fail to provide useful information regarding correlations between different pixels across a given image. In this regard, texture-based features are the most informative ones, specially for problems where tissue heterogeneity plays an important role, because texture-based features can catch the spatial relationships between neighboring pixels~\cite{Thawani:2017}. In Radiomics, typically,  texture-based features are extracted  based on different descriptive matrices, among them gray level co-occurrence matrix (GLCM), gray level run length matrix (GLRLM), neighborhood gray tone difference matrix (NGTDM), and gray level zone length matrix (GLZLM) are the most commonly used ones~\cite{Parekh:2016}, which are defined~below:

\begin{itemize}
\item \textbf{The GLCM}, models the spatial distribution of pixels' intensities and can be calculated by considering the frequency of the occurrence of all pairs of intensity values. Features extracted from GLCM are the most commonly used textural features in Radiomics~\cite{Parekh:2016}. Each GLCM is associated with two predefined parameters $\theta$ and $d$, where $\theta\in \{0^\circ,45^\circ, 90^\circ,135^\circ\}$, and $d$ is any integer distance admissible within the image dimensions.

\item \textbf{The GLRLM}, defines the number of adjacent pixels having the same intensity value, e.g., the $(i,j)$ element of the $GLRLM_\theta$ matrix determines the number of times intensity value $i$ has occurred with run length $j$, in direction $\theta$.

\item \textbf{The NGTDM}, which is based on visual characteristics  of the image, is a vector whose $k^{\text{th}}$ element is defined as the summation of differences between all pixels with intensity value $k$ and the average intensity of their neighborhood (size of which is determined by the user).
\vspace{-.01in}
\item \textbf{The GLZLM}, which looks for zones in a matrix. A zone can be defined as the set of connected pixels/voxels sharing the same intensity. The $(i,j)^{th}$ element of the GLZLM corresponds to the number of zones with the intensity $i$, and the size $j$.
\end{itemize}
%
\noindent
\textbf{4. Higher Order Radiomics Features}:
Higher order features such as Wavelet and Fourier features capture imaging bio-markers in various frequencies~\cite{Thawani:2017}. Wavelet features are the mostly used higher order features in Radiomics. Wavelet course and fine coefficients represent texture and gradient features respectively, and is calculated by multiplying the image by a matrix including complex linear or radial ``wavelet mother functions''. Fourier features can also capture gradient information. Minkowski Functional (MF) is another common higher order feature extractor considering the patterns of pixels with intensities above a predefined threshold.

In brief,  the MFs are computed by initially forming a binary version of the ROI through utilization of several thresholds within the minimum and maximum intensity limits. Although the number of utilized thresholds is a free parameter, for better results, it should be identified through a selection mechanism (typically empirical tests are used). Based on the binarized ROI, different MFs such as area and perimeter are computable~as follows
\begin{eqnarray}
MF_{\text{area}} &=& n_s,\\
\text{and } \quad MF_{\text{perimeter}} &=& -4n_s + 2n_e,
\end{eqnarray}
 where $n_s$ and $n_e$ are the total number of white pixels (above the threshold) and edges, respectively. This completes our coverage of feature extraction methods used in Radiomics.

\subsection{Radiomics Feature Reduction Techniques} \label{sec:RFR}
\begin{table*}[t!]
\small
\centering
\caption{Feature reduction techniques commonly used within the Radiomics literature.}
\begin{tabularx}{\textwidth}{XXX}
\arrayrulecolor{LightCyan}\hline
\rowcolor{LightCyan}
\textbf{~~~~~~~~~~~~~~~~~~Category} & \textbf{~~~~~~~~~~~~~~~~~~Description} & \textbf{~~~~~~~~~~~~~~~~~~Methods} \\

Supervised  &  Considers the relation of features with the class labels and features are selected mostly based on their contribution to distinguish classes. & \\
\\
~~$\bullet$ Filtering (Univariate) & Test the relation between the features and the class label one by one. &
Fisher score (FSCR); Wilcoxon rank sum test; Gini index (GINI); Mutual information feature selection (MIFS);  Minimum redundancy maximum relevance (MRMR), and; Student $t$-test~\cite{Parekh:2016}.\\
\\
~~$\bullet$  Wrapper (Multivariate)& Considers both relevancy and redundancy. & Greedy forward selection, and Greedy backward elimination. \\
\\
Unsupervised & Does not consider the class labels and its objective is to remove redundant features. & \\
\\
~~$\bullet$ Linear & Features have linear correlations. & Principle Component Analysis (PCA), and; Multidimensional scaling (MDS)\\
\\
~~$\bullet$ Nonlinear& Features are not assumed to be lied on a linear space. & Isometric mapping (Isomap), and; Locally linear embedding (LLE).\\
\arrayrulecolor{LightCyan}\midrule
\end{tabularx}
\label{tab:Reduc}
\end{table*}

Feature reduction is another critical step in Radiomics as although a large number of quantitative features can be extracted from the available image datasets, most of the features are highly correlated, irrelevant to the task at hand, and/or contribute to over-fitting of the model (making it highly sensitive to noise). Feature reduction techniques that are used in Radiomics can be classified into supervised and unsupervised categories~\cite{Zhang:2017}, as summarized in Table~\ref{tab:Reduc}. Supervised approaches, such as filtering and wrapper methods, take the discriminative ability of features into account and favor features that can best distinguish data based on a pre-defined class. Unsupervised methods, on the other hand, aim to reduce feature redundancy and include Principle Component Analysis (PCA), Independent Component Analysis (ICA) and Zero Variance (ZV)~\cite{Zhang:2017}.

In summery, various objectives can be defined when reducing the feature space in Radiomics. The following key characteristics can be defined for feature selection purposes \cite{Kumar:2012, Gillies:2016}:
\begin{itemize}
\item \textit{Reproducibility}: Reproducible features (also referred to as ``stable features'') are the ones that are more robust to pre-processing and manual annotations. These features will be discussed in Sub-section~\ref{sec:stab}.
\item \textit{Informativeness and Relevancy}, which can be defined as features that are highly associated with the target variable~\cite{Thawani:2017}. For instance a $\chi^2$-test, calculates the chi-squared statistic between features and the class variable, and consequently features with low impact on the target are discarded.  Another selection approach is a Fisher score test, where features with higher variance are treated as the more informative ones.
\item \textit{Redundancy}: Non-redundant features are the ones with small correlation with each other. Feature redundancy is defined as the amount of redundancy present in a particular feature with respect to the set of already selected features.
\end{itemize}
Below, supervised and unsupervised techniques commonly used in Radiomics are further discussed.

\vspace{.1in}
\noindent
\textbf{1. Supervised Feature Selection Methodologies}:
Supervised methods are generally divided into two categories as outlined below:
\begin{itemize}
\item \textit{Filtering (Univariate) Methods}: These methods consider the relation between the features and the class label one at a time without considering their redundancy. Among all filtering approaches, Wilcoxon test based method has been shown to be more stable, resulting in more promising predictions in the field of Radiomics~\cite{Parekh:2016}. A Wilcoxon test is a nonparametric statistical hypotheses testing technique that is used to determine dependencies of two different feature sets, i.e., whether or not they have the same probability distribution.

\item \textit{Wrapper (Multivariate) Methods}: Filtering methods have the drawback of ignoring relations between features which has led to development of wrapper techniques. In contrary to the filtering methods,  wrapper methods investigate the combined predictive performance of a subset of features, and the scoring is a weighted sum of both relevancy and redundancy~\cite{Parekh:2016}. However, computational difficulties prevent such methods from testing all the possible feature subsets.

Wrappers methods include greedy forward selection and greedy backward elimination. In a forward feature reduction path, selection begins with an empty set and the correlation with class label is calculated for all features individually. Consequently, the most correlated feature is selected and added to the set. In the next step, the remaining features are added, one by one, to this set to test the performance of the obtained set, and the process continues until no further addition can increase the predictive performance of the set. A backward selection path works in contrary to the forward one, beginning with a set including all the available features, and gradually reduces them until no further reduction improves the performance.
\end{itemize}
Since supervised methods are based on class labels, they are subject to over-fitting and can not be easily applied to different applications once trained based on a given feature set.

\vspace{.1in}
\noindent
\textbf{2. Unsupervised Feature Selection Methodologies}:
Unsupervised approaches try to reduce the feature space dimensionality by removing redundant features (those who are correlated and do not provide any additional information). Although these methods are not prone to over-fitting, they are not guaranteed to result in the optimum feature space. Unsupervised techniques can be divided into linear and non-linear methods, where the former assumes that features lie on a linear space. Because, in the field of Radiomics, commonly, very simple and basic forms of unsupervised techniques, such as PCA, are used, they are not covered in this article. However, this presents an opportunistic venue for application of more advanced statistical-based dimensionality reduction solutions recently developed within signal processing literature.

\subsection{Radiomics Statistical Analysis} \label{sec:RSA}
\begin{table*}[t!]
\small
\centering
\caption{Common analysis methods in Radiomics.}
\begin{tabularx}{\textwidth}{XXX}
\arrayrulecolor{LightCyan}\hline
\rowcolor{LightCyan}
\textbf{~~~~~~~~~~~~~~~~~~~~Purpose} & \textbf{~~~~~~~~~~~~~~~~~~~~Description} & \textbf{~~~~~~~~~~~~~~~~~~~~Methods} \\
Clustering & Similar patients are grouped together based on a distance metrics. &Hierarchical, Partitional \\
\\
Classification  & Models are trained to distinguish patients based on their associated clinical outcome. & Random Forest (RF);  Support Vector Machine (SVM); Neural Network (NN); Generalized linear model (GLM); Naive Bayes (NB); k-nearest neighbor (KNN); Mixture Discriminant Analysis (MDA); Partial Least Squares GLM (PLS), and; Decision Tree (DT).\\
\\
Time-related analysis & The survival time or the probability of survival is calculated based on the available set of data from previous patients.  &Kaplan-Meier survival analysis;  Cox proportional hazards regression model~\cite{Oikonomou:2018}, and; Log-Rank Test.\\
\arrayrulecolor{LightCyan}\midrule
\end{tabularx}
\label{tab:analys}
\end{table*}

Statistical analysis refers to utilizing the extracted Radiomics features in a specific task such as cancer diagnosis, tumor stage classification, and survivability analysis, as described in Section~\ref{sec:App}. Although most statistical methods, initially, treat all the features equally and use the same weights over all predictors, in the area of Radiomics, the most successful methods are the ones that use a prior assumption (provided by experts) over the meaning of features~\cite{Gillies:2016}. One basic approach to analyze the Radiomics features adopted in~\cite{Aerts:2014,Griethuysen:2017} is to cluster the extracted features and look for associations among clusters and clinical outcomes. For instance, patients belonging to one cluster may have similar diagnosis or patterns. Observations show that image bio-markers are associated with clinical outcomes such as tumor malignancy. Hierarchical clustering is most commonly used in Radiomics~\cite{Kumar:2012}. However, clustering techniques are not basically trained for target forecasting purposes, which necessitates the use of prediction tools that are specially trained based on a predefined class label. Prediction tools in Radiomics are categorized as either:
\begin{itemize}
\item[(i)]~\textit{Classification and Regression Models} that are mostly similar to other multi-media domains, trying to foresee a discrete or continues value. Random Forest (RF), Support Vector Machine (SVM) and Neural Network (NN) are among the most common regression and classification techniques used to make predictions based on Radiomics~\cite{Zhang:2017}.
\item[(ii)] \textit{Survivability analysis}: Also referred to as time-related models, mostly try to predict the survival time associated with patients. These models are also useful when testing the effectiveness of a new treatment.
\end{itemize}
Table~\ref{tab:analys} presents a summary of different Radiomics analysis techniques. As predictors belonging to the former category are also common in other multi-media applications, they are not covered in this article. Survivability analysis (the latter category), however, is more specific to Radiomics, and as such, below, we discuss the three mostly used techniques from this category, i.e., Kaplan-Meier Survival Curve (KMS), Cox Proportional Hazards (regression) Model (PHM), and Log-Rank Test.

\vspace{.1in}
\noindent
\textbf{\textit{1. Kaplan-Meier Survival Curve (KMS)}}:
The KMS curve~\cite{Aerts:2014, Zhang:2017} represents a trajectory for measuring the probability of survival $S(t)$ in given points of time $t$, i.e.,
\begin{equation}
\begin{aligned}
\centering
S(t)= &\frac{Number\ of\ patients\ survived\ until\ t}{Number\ of\ patients\ at\ the\ beginning}.
\end{aligned}
\end{equation}
The KMS curve can be calculated  for all Radiomics features to assess the impact of different features on patients' survival as follows:
\begin{enumerate}
\item A desired feature, for which the KMS curve is supposed to be calculated, is selected.
\item Based on the selected feature, one or more thresholds are considered that can partition patients into, e.g., low and high risk cancer subjects. Patients are then grouped based on whether their associated feature lies above or below the threshold.
\item The KMS curve is calculated for all the obtained groups, and the result can be used to compare the survivability among patients with, e.g., low and high risk cancer. For instance, in Reference~\cite{Aerts:2014} \textit{high heterogeneity features} are associated with shorter survival time, while \textit{high compactness features} are associated with longer survival.
\end{enumerate}

\vspace{.1in}
\noindent
\textbf{\textit{2. Cox Proportional Hazards (Regression) Model (PHM)}}~\cite{Aerts:2014},
is commonly used in medical areas to predict patient's survival time based on one or more predictors (referred to as covariates) such as Radiomics features. The output of the PHM model denoted by $h(t)$ is the risk of dying at a particular time $t$, which can be calculated as follows
\begin{equation}
h(t) = h_0(t)\times \exp^{\sum_{i=1}^{N_c}b_ix_i} \label{eq:PHM}
\end{equation}
where $x_i$, for ($1 \leq i \leq N_c$), are predictors (covariates); $b_i$ represent the impacts of predictors, and $h_0(t)$ is called the base-line hazard.  The exponent term in Eq.~\eqref{eq:PHM} is referred to as the ``Risk'' and  is conventionally assumed to be a linear combination of the features (covariates), i.e., $\text{Risk} \triangleq \sum_{i=1}^{N_c}b_ix_i$. The Risk coefficients ($b_i$, for ($1 \leq i \leq N_c$)) are then computed through a training process based on historical data. More realistically,  the risk can be modeled as a general non-linear function, i.e.,  $\text{Risk} \triangleq \f(\x)$, with the non-linearity being learned via deep learning architectures, which has not yet been investigated within the Radiomics context.

\vspace{.1in}
\noindent
\textbf{\textit{3. Log-Rank Test}}~\cite{Aerts:2014}, 
which is used for comparing the survival of two samples specially when these two samples have undergone different treatments. This test is a non-parametric hypothesis test assessing whether two survival curves vary significantly. One limitation associated with the Log-Rank test is that the size of the groups can influence the results,  therefore, larger number of patients should be included to from equal sized groups.

\noindent
\textbf{\textit{Evaluation of HCR}}:
In summary, having a successful hand-crafted Radiomics pipeline requires an accurate design to choose the best combination of feature extraction, feature reduction, and analysis methods. Several studies have tried to find the most important factor leading to the performance variation. For instance, the effects of these design choices are recently investigated in~\cite{Zhang:2017} through an analysis of variance, where it has been shown that, e.g., feature selection can significantly influence the final accuracy. On the other hand, References~\cite{CParmar:2015,PGrossmann:2015} have concluded that the classification method is the dominant source of performance variation for head-and-neck and lung cancer classification tasks. The difference between the concluding remarks of Reference~\cite{Zhang:2017} and References~\cite{CParmar:2015,PGrossmann:2015} indicates that impact of design choices may vary from one application to another, one dataset to another, and from a set of features to another.

Finally, it should be noted that reporting accuracy is not as informative measure in Radiomics as it is in other multi-media domains. Because, in medical areas, making mistakes in, for instance, classifying positive and negative samples are not equal. Therefore, in Radiomics, measures which are capable of distinguishing between False Positive (FP) and False Negative (FN) errors are more favored. Such one measure is the area under Receiver Operating Characteristic (ROC) curve, which allows for investigating the impact of different decision thresholds on FP and FN rates. Confusion matrix is another common and useful technique to report the performance of a Radiomics classifier in terms of its FP and FN rates.
In practice, most of the decisions in medical areas cannot be made with a complete certainty, and  physicians consider several factors such as harms and benefits of a specific judgment, when forming thresholds for their decisions. However, these factors are not quantified and utilized in Radiomics, which calls for a broad investigation into potential solutions to incorporate the factors commonly used by physicians.

\subsection{Radiomics Stability} \label{sec:stab}
An important aspect of Radiomics is the stability of the extracted features, which quantifies the degree of dependency between features and pre-processing steps.  Stability in Radiomics is generally evaluated based on either of the following two techniques:
\begin{enumerate}
\item \textbf{Test-Retest}: In this approach,  patients undergo an imaging exam more than once and images are collected separately. Radiomics features are then extracted from all the obtained sets and analyzed. Here, being invariant across different set of images illustrates stability of Radiomics features.
\item \textbf{Inter-observer reliability}, which is referred to an experiment where multiple observers are asked to delineate the ROI from the same images, and Radiomics features are extracted from all different delineations to test their stability for variation in segmentation~\cite{Griethuysen:2017}.
Here, being invariant across different segmentations  illustrates stability of Radiomics features.
\end{enumerate}
%
Different stability Criteria are used to find robust features in Radiomics as briefly outlined below:
\begin{enumerate}
\item \textbf{\textit{Intra-class Correlation Coefficient (ICC)}}: One approach to measure the stability of Radiomics features, which is used for both the aforementioned categories (i.e., test-retest and inter-observer setting) is referred to as the intra-class correlation coefficient (ICC)~\cite{Aerts:2014}. The ICC is defined as a metric of the reliability of features, taking values between $0$ and $1$, where $0$ means no reliability and $1$ indicates complete reliability. Defining terms $BMS$ and $WMS$ as  mean squares (measure of variance) between and within subjects, which are calculated based on a one-way Analysis of variance (ANOVA), for a test-retest setting, the ICC can be estimated as
\begin{equation}
\centering
ICC_{\text{Test-Retest}}=\frac{BMS-WMS}{BMS+(N-1)WMS},
\end{equation}
where $N$ is the number of repeated examinations. By defining $EMS$ as residual mean squares from a two-way ANOVA and $M$ as the number of observers, for an inter-observer setting, the ICC can be calculated~as
\begin{equation}
\centering
ICC_{\text{Inter-Observer}}=\frac{BMS-EMS}{BMS+(M-1)EMS}. \label{eq:6}
\end{equation}
\item \textbf{\textit{Friedman Test}}: The Friedman test, which is specially useful for assessing the stability in an inter-observer setting, is a nonparametric repeated measurement that estimates whether there is a significant difference between the distribution of multiple observations, and has the advantage of not requiring a Gaussian population. Based on this test, the most stable features are the ones with a stability rank of $1$ \cite{Aerts:2014}.

\end{enumerate}
In~\cite{Aerts:2014}, it is declared that Radiomics features with higher stability have more prognostic performance,  therefore, stability analysis can be interpreted as a feature reduction technique. According to Reference~\cite{Griethuysen:2017}, Laplacian of Gaussian (LoG), intensity-based, and texture features are more stable for lung CT images, while wavelet and shape-based features are sensitive to variation in segmentation. However, there are also other sources of variation (other than the segmentation step), that can influence the stability of the features, one of which is the image intensity discretization strategy~\cite{Leijenaar:2015} that has a strong impact on the texture features, in particular. There are two main approaches to discretize the medical images. The first one is to adopt a fixed bin size for all of the images, and the second one is to use a fixed number of bins. While it is shown that both approaches lead to texture features that depend on the intensity resolution, the first method (fixed-sized bins) results in more stable and comparable features. Nevertheless, texture analysis requires standardized intensity discretization method, to serve as a meaningful and reliable Radiomics technique.

Finally, it is worth mentioning that the Image Bio-marker Standardization Initiative (IBSI)~\cite{Zwanenburg:2018} is an international collaboration that seeks to provide unique definitions, guidelines, and Radiomics steps, in response to the reproducibility challenge of the Radiomics. The provided guideline covers several steps within the Radiomics pipeline, from image acquisition, pre-processing and segmentation, to feature calculation. To summarize our finding on $\HCR$ and elaborate on its applications, we have provided an example in Information Post \ref{infoPst2}, where the problem of Radiomics-based lung cancer analysis is investigated.

\subsection{Radiogenomics}\label{Sec:RG}

Radiomics is typically combined with Genomic data, often referred to as Radiogenomics~\cite{Gillies:2016}. In other words, Radiogenomics refers to the relationship between the imaging characteristics of a disease and its gene expression patterns, gene mutations and other genome-related characteristics. Potential association between imaging outcome and molecular diagnostic data can serve as a predictor of patient's response to the treatments and provide critical support for the decision making tasks within the clinical care setting. In other words, Radiogenomics has the potential to investigate the Genomics of  the cancer, without utilizing invasive procedures such as biopsy. Various association mining and clustering methods are used to identify the relationships between gene expressions and Radiomics, e.g., in~\cite{Gillies:2016} it was found that just $28$ Radiomics features were able to reconstruct $78\%$ of the global gene expressions in human liver cancer cells.

\begin{figure*}[t!]
\centering
\includegraphics[width=1\textwidth]{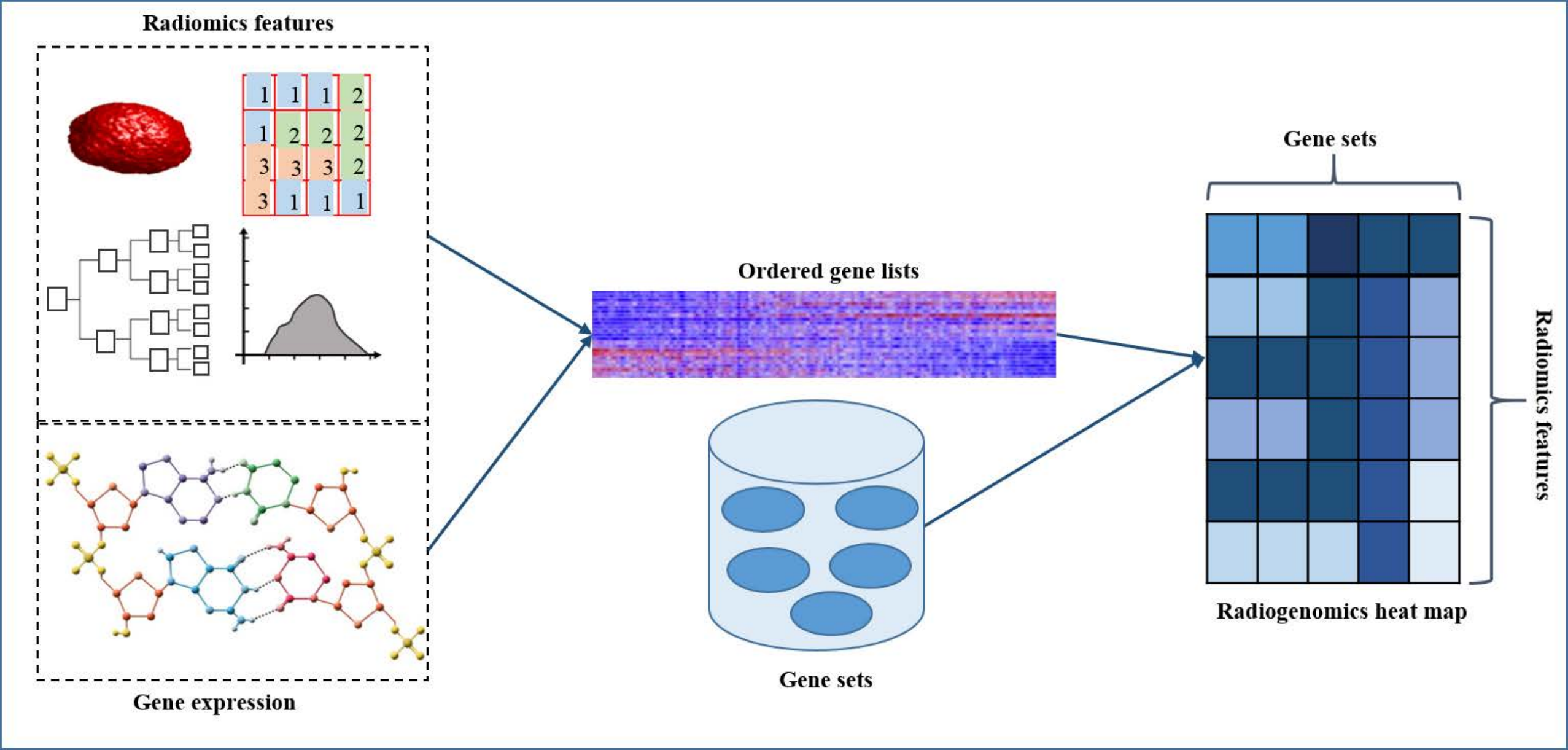}
 \centering
\caption{Radiogenomics analysis using GSEA. Genes are first sorted based on their association with Radiomics features, and the final heat map is formed according to the pattern of gene sets within the ordered gene list. }
\label{fig:RaGe}
\end{figure*}

To assess the association between gene expression and discrete classes such as benign and malignant tumors, genes should be first sorted based on their discriminative ability. However, the goal of Radiogenomics is to find associations between gene expression and Radiomics features, therefore, discriminative ability is indefinable. Spearman's Rank Correlation Coefficient (SRC) \cite{Parmar:2017}  can be used to measure the correlation between a specific Radiomics feature and gene expression. Genes are then sorted based on their SRC coefficient instead of their discriminative ability. The ordered genes are typically stored in a list \textit{L}, and to extract meaning from this list, the traditional approach is to focus on the top and bottom genes in list \textit{L}, representing genes with the strongest positive and negative correlations, respectively. This approach, however, is subject to several limitations, such as the difficulty in biological interpretation, which has led to the introduction of Gene-set Enrichment Analysis (GSEA)~\cite{Subramanian:2005}. The GSEA is one of the mostly used Radiogenomics approaches \cite{Aerts:2014}, and is based on the definition of gene sets. Each gene set \textit{S} is a group of genes that are similar in terms of the prior biological knowledge, such as involvement in common biological pathways. The goal of the GSEA is to find out whether the members of a given set \textit{S} tend to occur in the top or bottom of the list \textit{L}. In this case, the expression of this gene set is associated with the specific Radiomics feature. The result of Radiogenomics analysis using GSEA is a heat map representing the degree of association between all gene sets and Radiomics features as shown in Fig.~\ref{fig:RaGe}.

The main role of Radiogenomics is that it allows imperfect data sets, where clinical outcomes are difficult to be collected or require an extended period to be collected, to be leveraged based on prior knowledge of the relationship between clinical outcomes and genomics, in order to draw new conclusions. For example in a research project there may have been imaging data and genome-related data but no clinical outcomes. If there is prior knowledge of the relationship of the genomics with certain clinical outcomes, then by correlating the imaging data with the genomics, new conclusion can be drawn about the relationship of the imaging data with the clinical outcomes. In this scenario genomics can fill the gaps in knowledge.

\begin{strip}
\begin{tcolorbox}[colback=green!5,colframe=green!40!black,title=Information Post II: Radiomics for Lung Cancer Analysis,label=infoPst2]
\vspace{-.1in}
\begin{center}
\includegraphics[width=11cm]{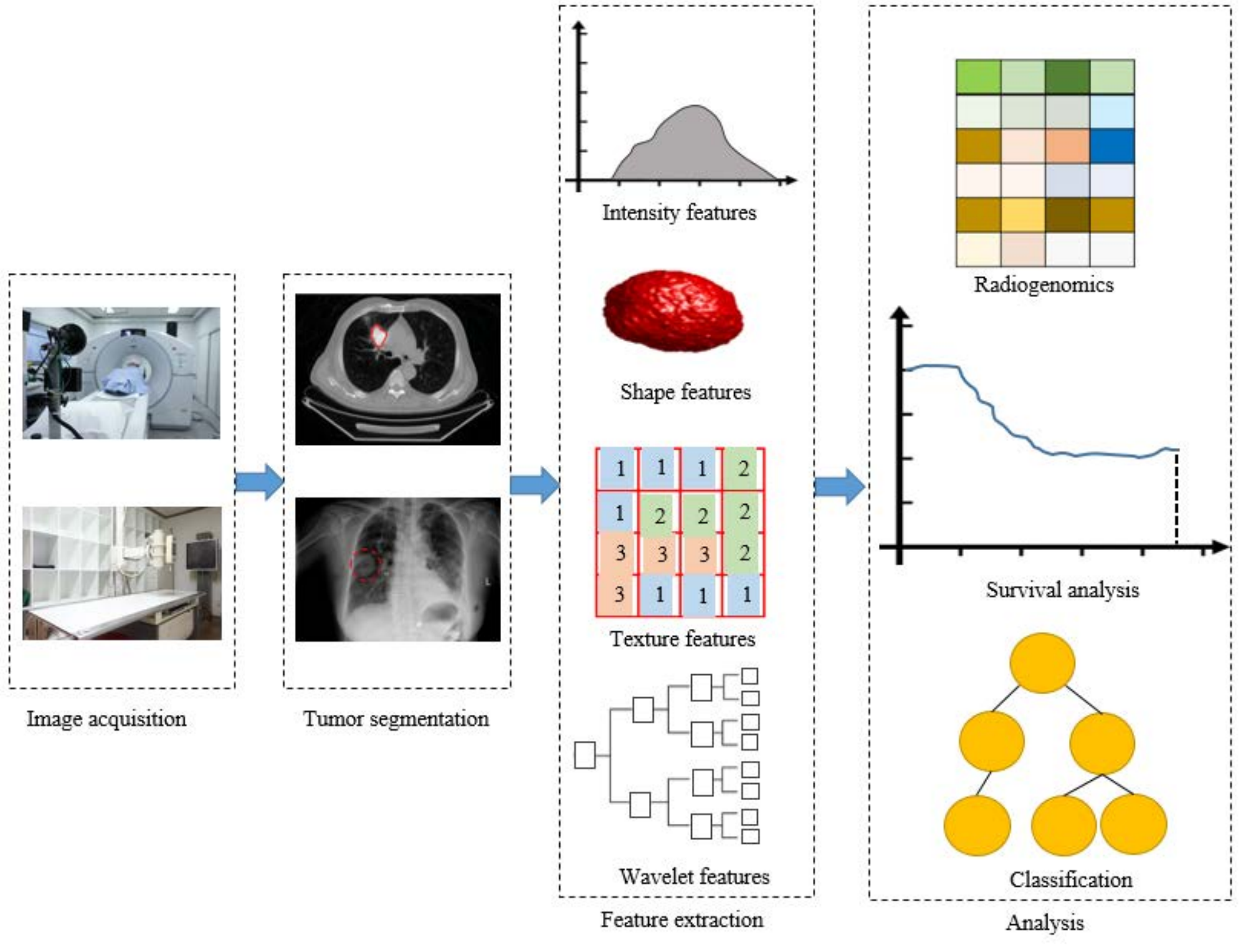}
\end{center}
\vspace{-.2in}
\begin{multicols}{2}
\small
Studies have shown that use of computer aided systems can significantly help with the early diagnosis and detection of lung cancer and consequently lead to dramatic decrease in lung cancer-related mortality \cite{Kumar:2015}. For these computer aided systems to be successful in lung cancer early prognosis, distinguishing Radiomics features have to be extracted from the CT images, which are the main detectors of lung cancer.
Radiomics can help with different tasks related to lung cancer such as: (i) Lung cancer patient survival prediction; (ii) Lung cancer malignancy prediction; (iii) Forecasting the patient's response to treatment; and (iv) Prediction of the stage of the cancer.

As illustrated in the figure above, the first stage in lung cancer Radiomics pipeline is commonly the segmentation of left and right lungs, followed by the \textbf{segmentation of the tumor} which can be performed automatically for tumors with high intensities located in low intensity backgrounds. However, segmentation can become problematic when lung tumors are attached to vessels or walls which makes automatic segmentation fail to generate reproducible and accurate outcomes. Due to this reason lung tumor segmentation is still an open problem~\cite{Kumar:2012}.

The next step after segmenting the tumor, is to \textbf{extract Radiomics features}. Types of features to extract depend on the problem at hand. For instance, according to \cite{Thawani:2017}, intensity features are highly correlated to lung cancer patient survival, while shape features are beneficial for lung cancer malignancy prediction and also forecasting the patient's response to treatment. More importantly, texture features are reported to be the most influential ones on lung cancer outcomes~\cite{Gillies:2016}. Selected features will then go through a \textbf{feature selection} process, where redundant and non-relevant features are excluded. Feature selection is followed by a \textbf{statistical analysis} step to perform the aforementioned tasks.

To further elaborate on the use of Radiomics in lung cancer analysis, we have implemented the Hand-Crafted pipeline on $157$ patients, from NSCLC-Radiomics dataset~\cite{Aerts:2014}. We grouped these patients into two categories: Long-term survival, referring to patients who have survived more then $500$ days, and Short term survival, referring to those who have survived less than $500$ days. The following techniques (from possible options described in Section III) are employed in the proposed pipeline:
\begin{enumerate}
\item \textbf{Segmentation}: Manual annotations performed by an expert are utilized.
\item \textbf{Radiomics Feature Extraction}: $11$ first-order (Number of pixels, ROI size in $mm^2$, Mean gray level, Standard Deviation, Median gray level, Min ROI, Max ROI, Mean Positive Values, Uniformity, Kurtosis, Skewness) and $19$ second-order Radiomics features (Contrast, Energy, Correlation, Homogeneity, Entropy, Normalized Entropy, Variance, Inverse Difference Moment, Sum of Average, Sum of Variance, Sum of Entropy, Difference of Variance, Difference of Entropy, Information Measure of Correlation, Autocorrelation, Dissimilarity, Cluster Shade, Cluster Prominence, Maximum Probability) are extracted from lung tumors~\cite{Zhang:2017}.
\item \textbf{Feature Reduction}: Based on a $\chi^2$ feature selection test, among all the extracted features, the surface area and volume of the tumors are selected as the most correlated ones to the survival outcome.
\item \textbf{Analysis}: Following figure illustrates the relation between survival and the two aforementioned features. As it can be inferred from this figure, although patients with short-term survival can be associated with various tumor sizes, those who have survived longer have smaller tumors in terms of surface area and volume. In other words, being small in size seems to be necessary for long-term survival, however, as so many other factors influence survival, not all small tumors lead to long survivals. For instance, the location of the tumor can have noticeable impact on survival, nevertheless, it can not be assessed based on hand-crafted Radiomics features, as they are extracted from the segmented ROI, without taking its location into account.
\begin{center}
\includegraphics[width=7cm]{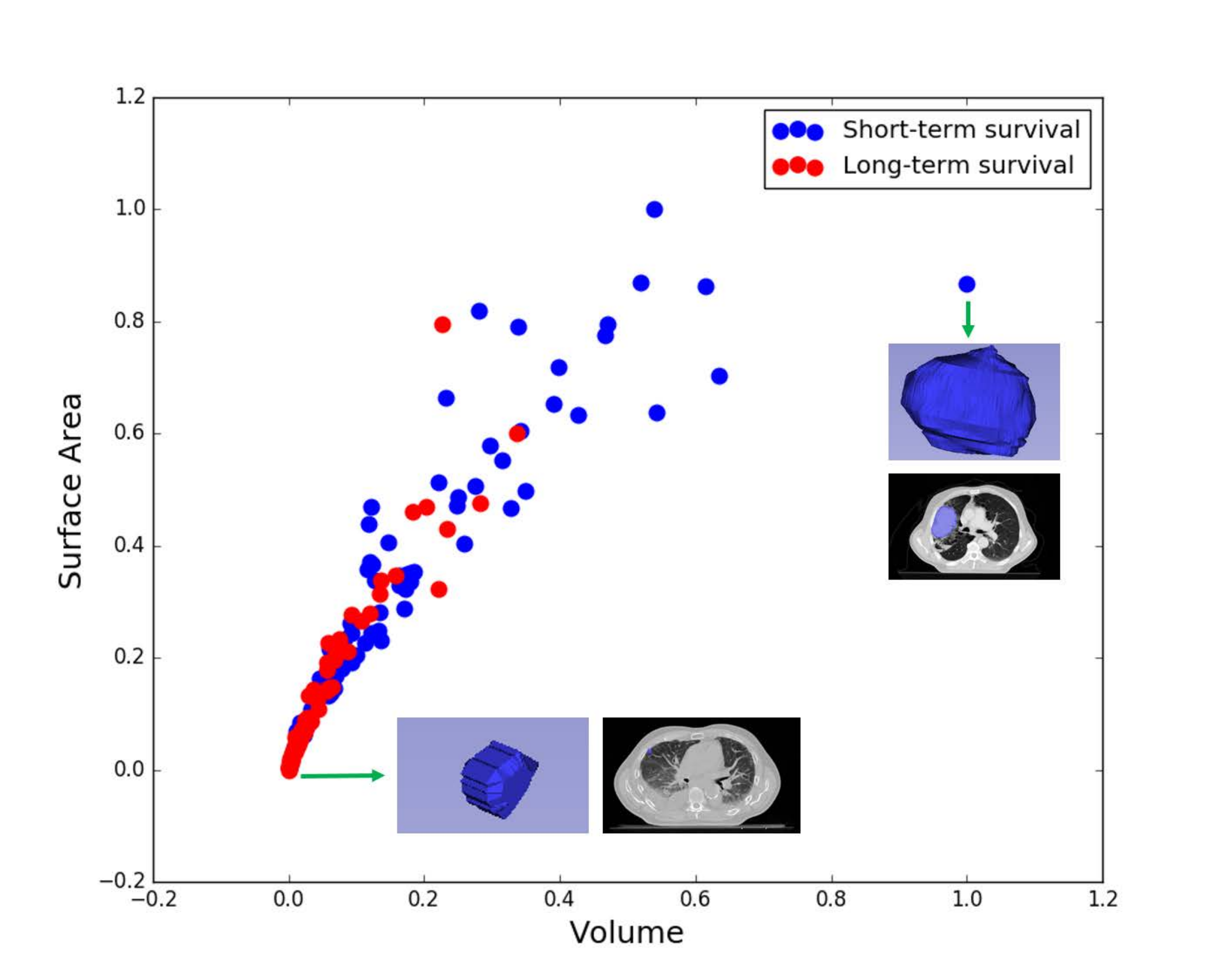}
\end{center}
\end{enumerate}

\vspace{-.2in}
\end{multicols}

\end{tcolorbox}

\end{strip}

\section{State-of-the-Art in Deep Learning-based Radiomics} \label{sec:DL_radiomics}
\begin{figure*}[t!]
\centering
\includegraphics[width=1\textwidth]{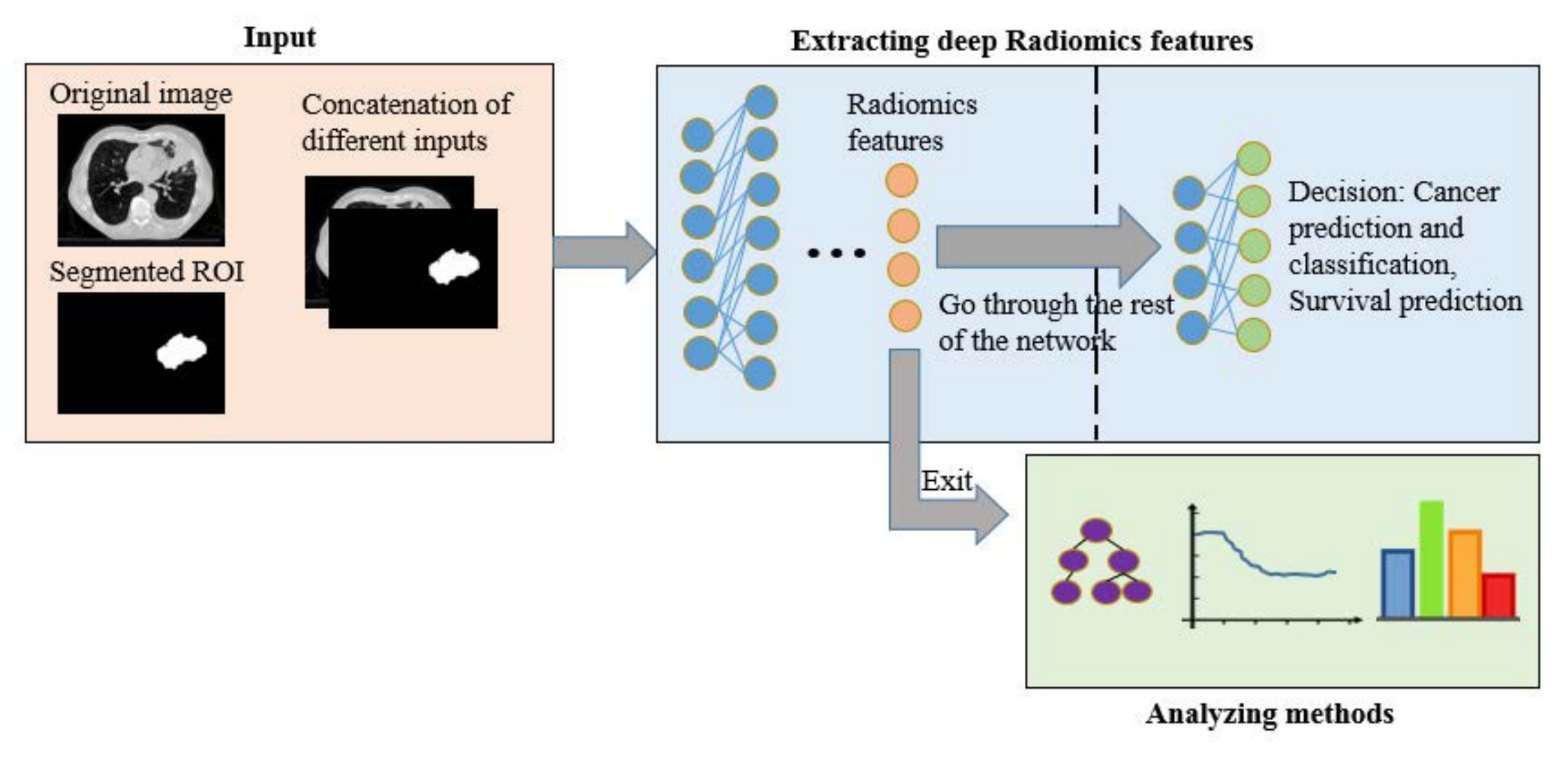}
\centering
\caption{\small Extracting deep Radiomics. The input to the network can be the original image, the segmented ROI, or the combination of both. Extracted Radiomics features are either utilized through the rest of the network, or an external model is used to make the decision based on Radiomics.}
\label{fig:Deep}
\end{figure*}

Deep learning-based Radiomics (DLR), sometimes referred to as ``Discovery Radiomics'' or ``Radiomics Sequence Discovery" with ``Sequence" referring to features~\cite{Shafiee:2017}, is the process of extracting deep features from medical images based on the specifications of a pre-defined task including but not limited to disease diagnostics;  Cancer type prediction, and; Survival Prediction.
In brief, the $\DLR$ can be extracted via different architectures (stack of linear and non-linear functions), e.g.,  Convolutional Neural Network (CNN) or an Auto-Encoder, to  find the most relevant features from the input~\cite{Kumar:2015}. Fig.~\ref{fig:Deep} illustrates the schematic of extracting deep features. The extracted features can then either go through the rest of the deep net for analysis and making decisions or they may exit the network and go through a different analyzer such as an SVM or a Decision Tree (DT). Commonly used deep architectures for Radiomics will be discussed in details later in Section~\ref{sec:DLArc}.

\noindent
 \textbf{\textit{Benefits of $\DLR$ vs. $\HCR$}}:
An important advantage of $\DLR$ over its hand-crafted counterpart is that the former does not need any prior knowledge and features can be extracted in a completely automatic fashion with high level features extracted from low level ones~\cite{Kumar:2015}. Moreover, deep learning networks can be trained in a simple end-to-end process, and their performance can be improved systematically as they are fed with more training samples \cite{Cheng:2016}. Another key benefit of using $\DLR$ instead of $\HCR$ is that the input to the deep networks to extract Radiomics features, can be the raw image without segmenting the region of interest, which serves the process in two ways:
\begin{enumerate}
\item[(i)] Eliminating the segmentation step can significantly reduce the computational time and cost by taking the burden of manual delineation off the experts and radiologists, besides, manual annotations are highly observer-dependent, which makes them unreliable sources of information, and;
\item [(ii)] Automatic segmentation methods are still highly error prone and inaccurate to be used in a sensitive decision making process.
\end{enumerate}
Furthermore, the input to a deep network can also be the combination of the original and segmented image along with any other pre-processed input such as the gradient image (referred to as ``multi-channel'' input), all concatenated along the third dimension~\cite{Sun:2017}. The variety of input types can even go further to include images from different angles such as coronal and axial~\cite{Ciompi:2017}.

Generally speaking, studies on $\DLR$ can be categorized from several aspects including:
\begin{enumerate}
\item[(i)] \textit{Input Hierarchy}: The input to the deep net can be the single slices, the whole volume, or even the whole examinations associated with a specific patient. Each of these cases require their own strategy, e.g., in case of processing the whole volume simultaneously, one should think of a way to deal with the inconsistent dimension size, as patients are associated with different number of slices. One common architecture that allows for utilization of inputs with variable sizes, such as various number of slices, is the Recurrent Neural Network (RNN), which will be briefly discussed in Section~\ref{sec:DLArc};
\item[(ii)] \textit{Pre-trained and Raw Models}: Depending on the size of the available dataset and also the allocatable time, pre-trained models can be fine-tuned or raw models can be trained from scratch. This will be analyzed more specifically in Section~\ref{subsec:PtrRaw}, and;
\item[(iii)] \textit{Deep Learning Network Architectures}: Choice of the deep network is the most important decision one should make to extract meaningful and practical $\DLR$, which will be discussed in Section~\ref{sec:DLArc}.
\end{enumerate}
%
\begin{figure*}[t!]
\centering
\includegraphics[width=1\textwidth]{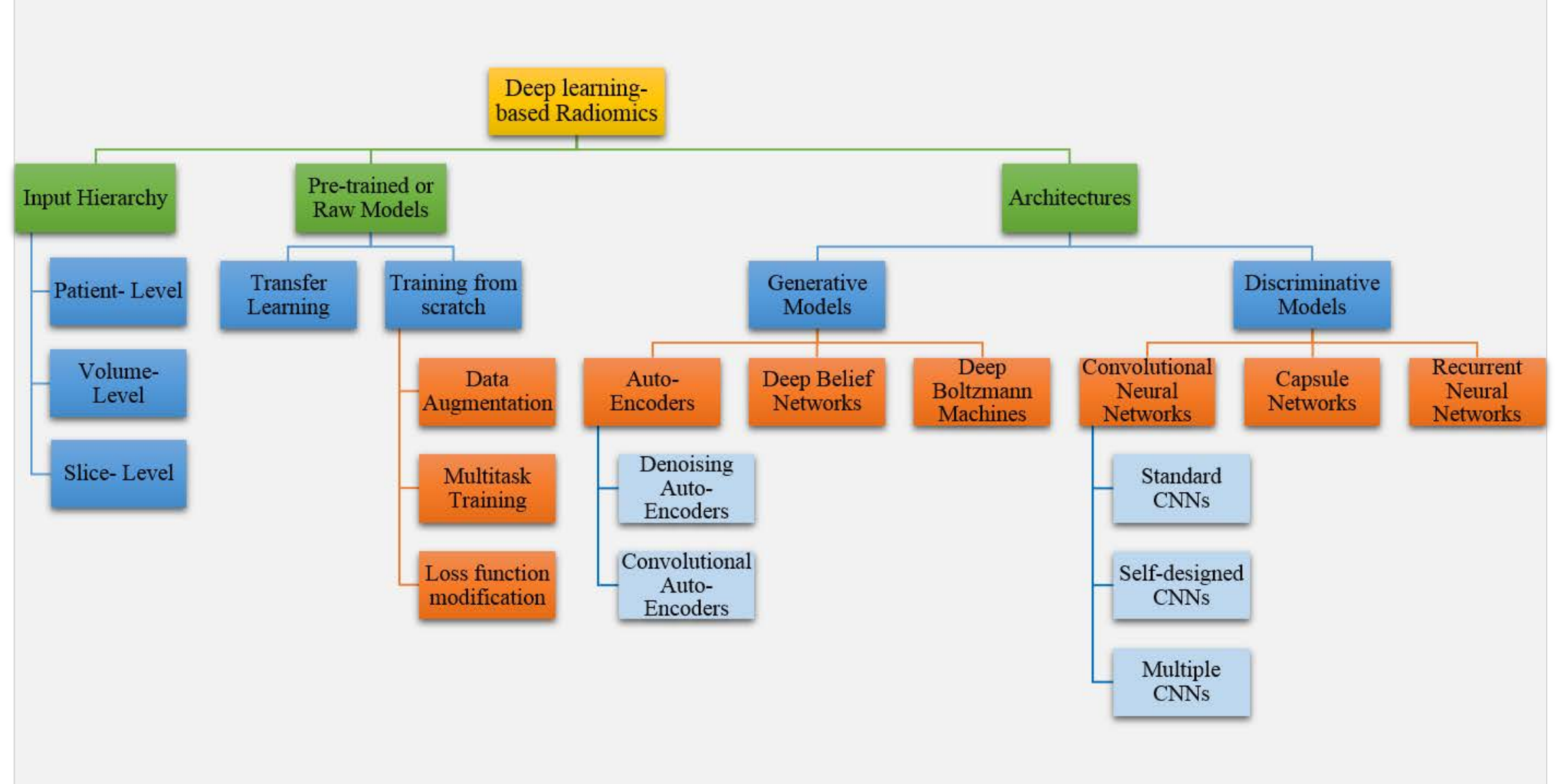}
 \centering
\caption{\small Taxonomy of Deep Learning-based Radiomics (DLR).}
\label{fig:map}
\end{figure*}
In the reminder of this section, we will review the state-of-the-art in deep Radiomics from different perspectives such as input hierarchy, pre-trained vs. raw models, and deep learning network architectures. Fig.~\ref{fig:map} illustrates a taxonomy of different DLR approaches providing a guide to the rest of this section.

\subsection{Input Hierarchy}\label{subsec:InpHir}
\begin{figure*}[t!]
\centering
\includegraphics[width=1\textwidth]{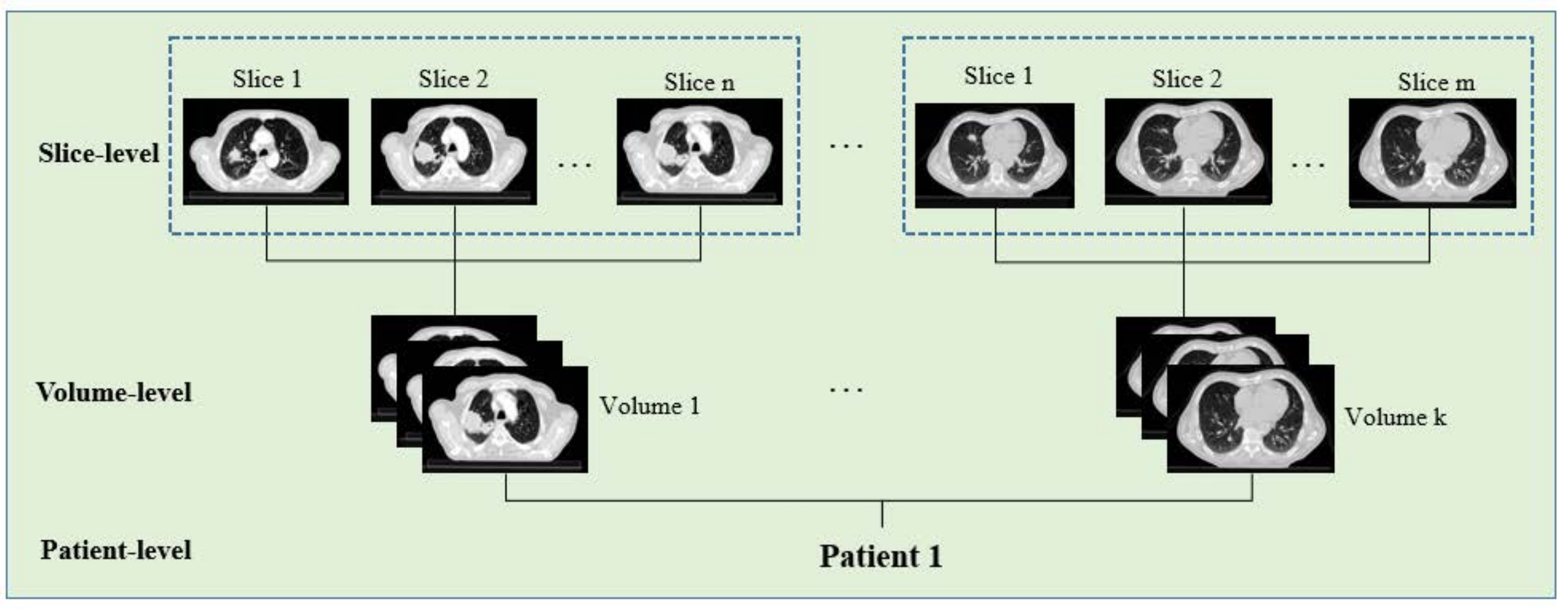}
 \centering
\caption{\small Input hierarchy for one patient. In the top row the slice-level input is shown where the patients went through $K$ examination visits during each of which $N\i$, for ($1 \leq i \leq K$), number of slices is captured. The second row shows the volume-level where all slices associated with one visit is provided simultaneously as the input to the network. Finally, the third row shows the Patient-level analysis, where a single input consisting of all the volumes is provided.}
\label{fig:level}
\end{figure*}
As shown in Fig.~\ref{fig:level}, input images for $\DLR$ studies can be divided into three main categories: Slice-level; Volume-level, and; Patient-level. Slice-level classification refers to analyzing and classifying image slices independent from each other, however, this approach is not informative enough as we typically need to make decisions based on the labels assigned to the entire Volume of Interest (VOI). Shortcomings of slice-level classification leads to another approach referred to as volume-level classification, where either  the slice-level outputs are fused through a voting system, or the entire image slices associated with a volume is used as the input to the classifier. Finally, patient-level classification refers to assigning a label to a patient based on a series of studies (such as CT imaging follow-ups). For example, in Reference~\cite{WShen:2016}, patient-level classification  is explored  with the goal of  estimating the probability of lung tumor malignancy based on a set of CT studies. To achieve this goal, initially, a simple three layer CNN is trained to extract $\DLR$ from tumor patches associated with individual CT series (volume-level classification) with the objective of minimizing the difference between the predicted malignancy rate and the actual rate. Then, by adopting a previously trained CNN, the malignancy rate is calculated for all the series belonging to the patient and the final decision is made by selecting the maximum malignancy rate. In other words, a patient is diagnosed with malignant lung cancer if at least one of the predicted rates is above a pre-determined rate for malignancy.

\subsection{Pre-trained or Raw Models}\label{subsec:PtrRaw}
Similar to the other medical areas, the $\DLR$ can be extracted based on either of the following two approaches:

\vspace{.1in}
\noindent
\textbf{\textit{Training from scratch}}: Training a deep network \textcolor{black}from scratch for extracting $\DLR$ has the advantage of having a network completely adjusted to the specific problem at hand. However, performance of training from scratch could be limited due to couple of key issues, i.e., over-fitting and class imbalance. Adhering to patients' privacy and need for experts to provide ground truth typically limits the amount of  medical datasets available for extracting $\DLR$ resulting in over-fitting of the deep nets. The second issue is the problem of class imbalance, i.e., unequal number of positive and negative classes. This happens as number of patients diagnosed with abnormalities is commonly less that the amount of data available from healthy subjects. More specifically, class imbalance in medical areas is due to the fact that typically number of positive labels is less than the number of negative ones, making the classifier biased toward the negative class, which is more harmful than the other way around because, for instance, classifying a cancerous patient (positive label) as healthy (negative label) has worse consequences than classifying a healthy patient as cancerous \cite{Echaniz:2017}. The following strategies can be adopted to address these two issues:
\begin{itemize}
\item[(i)] \textbf{\textit{Data Augmentation}}, where different spatial deformations (such as rotation~\cite{Chung:2017}) are applied to the existing data in order to generate new samples for training purposes. Sub-patch Expansion~\cite{Sun:2017} is another form of augmentation commonly adopted in Radiomics to handle the inadequate data situation via extracting several random fixed-sized sub-patches from the original images.
\item[(iii)] \textbf{\textit{Multitask training}} is another method introduced to handle class imbalance and inadequate data~\cite{Ravi:2017}, which is achieved by decreasing the number of free parameters and consequently the risk of over-fitting. For instance, this approach is adopted in~\cite{Jamaludin:2016} for spinal abnormality classification based on MRIs through training a multitask CNN. Multitask in this context refers to performing different classification tasks simultaneously  via the same unified network (e.g., the network tries to classify disk grading and disk narrowing at the same time). The loss function is defined as the weighted summation of all the losses associated with different tasks. One important decision to make in multitask learning is the point that branching begins, e.g., in Reference~\cite{Jamaludin:2016}, a unified CNN is trained, where all Convolutional layers are shared for performing different tasks  and tasks are separated from the point that fully connected layers begin.
\item[(iv)] \textbf{\textit{Loss function modification}}: Another common approach specific to handling class imbalance for $\DLR$ extraction is to modify the loss functions by giving more weight to the minority class~\cite{Jamaludin:2016}.
\end{itemize}
\vspace{.1in}
\noindent
\textbf{\textit{Transfer Learning via a Pre-trained Network}}: A different solution to class imbalance and inadequate training data is ``transfer learning"  followed by ``fine tuning"~\cite{Ravi:2017,Li:2016,Sun:2017}. The transfer learning phase refers to training the deep net using a natural image data set, and then in the fine tuning phase, the trained network will be re-trained using the desired medical dataset. This strategy is adopted in Reference~\cite{Li:2016}, where a pre-trained CNN is used for breast cancer classification based on mammographic images. The pre-trained CNN used is an Alexnet which is too complicated and prone to over-fitting for small datasets. Therefore, this network is first pre-trained using ImageNet database which consists of more than one million natural images. Pre-trained CNN based on ImageNet is also adopted in~\cite{Hawkins:2016} for lung cancer survival prediction.

\subsection{Deep Learning Architectures in Radiomics}\label{sec:DLArc}
Radiomics features can be extracted through both discriminative and/or generative deep learning networks. As is evident from its name, discriminative deep models try to extract features that make the classes (e.g., normal or cancerous) distinguishable, and thus these models can directly classify instances from the extracted features. On the other hand, generative models are unsupervised, meaning that they are trained without considering the class labels. Generally, the goal of these models is to learn the data distribution in a way that enables them to generate new data from the same distribution. In other words, generative models can extract the natural and representative features of the data, which can then be used as inputs to a classifier. Furthermore, in the field of Radiomics, it is common~\cite{Cheng:2016} to train a generative model and use the learned weights as initial weights of a discriminative model. Below, an introduction to widely used discriminative and generative deep models in Radiomics is provided.

\vspace{.1in}
\noindent
\textbf{1. Discriminative Models}:
Deep discriminative models try to extract features capable of distinguishing class labels, and the objective is to minimize the prediction error. Below, we will review the Convolutional Neural Networks (CNNs) and Recurrent Neural Networks (RNNs), which are the most  popular discriminative architectures in Radiomics. Later, we will introduce a recently designed deep architecture referred to as the Capsule network (CapsNet)~\cite{Hinton:2017} and explain how this new architecture can contribute to the Radiomics.
%

\vspace{.1in}
\noindent
\textit{\textbf{1.1. Convolutional Neural Networks (CNNs)}}:
CNN is a stack of layers performing Convolutional filtering combined with nonlinear activation functions and pooling layers~\cite{Ravi:2017}. The fact that CNNs have recently resulted in promising outcomes have made them the mostly used architecture in medical areas including Radiomics. CNNs are more practical in the sense that  shared weights are utilized over the entire input, which reduces the number of trainable parameters. Unlike extracting hand-crafted features, kernels used in Convolutional layers are not pre-determined and are automatically  learned through the training process. This property makes CNNs  suitable methods for extracting $\DLR$ features as they are flexible and can be applied without requiring a prior knowledge. In~\cite{WShen:2017}, it has been shown that the $\DLR$ extracted from a CNN can visually distinguish benign and malignant lung tumors when projected into a 2D space, while the original pixel values completely fail to provide such distinction.

When adopting CNNs in the field of Radiomics, output of the fully connected layers is typically treated as $\DLR$ features. These features are then either used within the original CNN to provide the desired (classification and/or regression) output such as cancer type, or exist the network to be provided as the input to the rest of the Radiomics pipeline. As an example, in Reference~\cite{Hawkins:2016} $\DLR$ are extracted from the layer just before the classification (SoftMax) layer of the CNN with the goal of lung cancer survival prediction. These features are referred to as the ``preReLU'' and ``postReLU'' features as they are extracted both before and after applying the ReLU activation function. The $\DLR$ features are then used as inputs to four classifiers (i.e., Naive Bayes, Nearest Neighbor, Decision tree and Random Forest) after going through a feature selection algorithm.

The CNN architectures used in Radiomics can be divided into three main categories: (i) Standard architectures; (ii) Self-designed architectures, and; (iii) Multiple CNNs. Below, we describe each of these categories with examples from Radiomics:
\begin{itemize}
\vspace{.1in}
\item[1.1.1.] \textbf{\textit{Standard CNN Architectures}}:  As the name suggests, standard architectures are those that have been previously designed to solve a specific problem, and due to their success are now being adopted in the Radiomics. Two of such architectures that have been used in Radiomics are LeNet and AlexNet. The LeNet is one of the simplest CNN architectures, having a total of 7 layers, that has been used in Radiomics. However, researchers have some times modified this network to achieve higher performance, e.g., the CNN used in Reference~\cite{Sun:2017} is a LeNet architecture with a total of 9 layers including 3 Convolutional layers, 3 pooling layers and one fully connected layer followed by the classification layer to classify lung tumors as either benign or malignant.

\vspace{.05in}
Another commonly used standard architecture in Radiomics is the $11$ layers CNN called Alexnet, which has been adopted in~\cite{Li:2016} to extract $\DLR$ features from breast mammographic images. Features are extracted from all $11$ layers and used as inputs to $11$ support vector machine (SVM) with the goal of classifying breast tumors as either benign or malignant. Since it is not obvious which set (output of which of the 11 underlying layers) of $\DLR$ features are more practical, these SVMs are compared and the one with the largest area under the curve is chosen  for predictive analysis of breast cancer. The results of~\cite{Li:2016} concluded that the features extracted from the $9^{\text{th}}$ layer (a fully connected layer before the last fully connected layer and the classification layer) are the best predictors of breast cancer and they are of lower dimension compared to previous ones, which reduces the computational cost. In other words and in contrary to~\cite{Hawkins:2016}, the output of the last Convolutional layer, right before the fully connected layer, is selected as the $\DLR$ features.

\vspace{.05in}
Although AlexNet is a powerful network, it has too many parameters for small datasets and is, therefore, prone to over-fitting. As a result, Reference~\cite{Zhou:2017} has used an Alexnet with number of layers reduced to $5$ in order to avoid the over-fitting problem. The input to this network is a combination of CT and PET images, each having $3$ channels: One slice corresponding to the center of the lung nodule, specified by an expert, and the two immediate neighbors. The goal of this article is to classify lung tumors as benign or malignant, and although it has been shown that the adopted CNN does not result in significantly higher accuracy than classical methods ($\HCR$), it is more convenient as it does not require the segmented ROI.

Inception network~\cite{Szegedy:2015,CSzegedy:2015} is another CNN adopted in Radiomics. This network involves parallel convolutions with different kernel sizes, and poolings within the same layer, with the overall aim of allowing the network to learn the best weights and select the most useful features. The Inception CNN is used in~\cite{Gulshan:2016}, for the detection of diabetic retinopathy. This paper is the first work on deep learning-based detection of diabetic retinopathy that has been approved by the Food and Drug Administration (FDA).

\vspace{.1in}
\item[1.1.2.] \textbf{\textit{Self-designed CNNs}}: As opposed to researchers that have used standard CNNs with or without modifications, some have designed their own architectures based on the specification of the Radiomics problem at hand. For example, Reference~\cite{Chung:2017} has used a CNN with three Convolutional layers to extract $\DLR$ features, and although the CNN itself is trained to use these features for classifying benign and malignant tumors, they are used as inputs to a binary decision tree.

\vspace{.05in}
In a similar fashion, Reference~\cite{Li:2017} has used a CNN with 6 Convolutional layers and one fully connected layer for $\DLR$ extraction in the problem of brain tumor classification. The designed network, however, is different from previously mentioned articles as it is developed for tumor segmentation, and \textit{features are extracted from the last Convolutional layer since they are more robust to shifting and scaling of the input}. In other words, the CNN was designed for segmentation and once trained, the output of the last Convolutional layer is used as the $\DLR$ features. The claim here is that the quality of extracted features depends on the accuracy of segmentation, and when segmentation is precise the quality of Radiomics features is guaranteed. Due to the high importance of the segmentation, more advanced and efficient CNN architectures have been developed, one of which is the Fully Convolutional Neural Network (FCNN)~\cite{Litjens:2017}. In an FCNN, fully connected layers are rewritten as convolutional layers, having the advantage of not requiring fixed-sized inputs. This network is also extended to 3D image segmentation, to segment multiple targets at once. To decrease the false positive rate, FCNN is further combined with graphical models such as Markov Random Fields (MRFs) and Conditional Random Fields (CRFs). Finally, to improve the resolution of the output, U-Nets~\cite{Ronneberger:2015} are proposed, which include up-convolutions to increase the image size, and skip-connections to recover spatial information.

\vspace{.05in}
Lung cancer detection using CNNs is also investigated in~\cite{Fu:2017}, with the difference that the input to the network is not only the original image but also the nodule-enhanced and vessel-enhanced images, stating that providing the network with more information on tumor and vessels reduces the risk of misplacing these two by the network. The main focus here is to reduce the false positive rate while keeping the sensitivity high, therefore, a significant number of nodule candidates are selected at the beginning. Use of CNNs is further investigated in~\cite{Oakden-Rayner:2017}, where a $7$ layer architecture is fed with down-sampled volumetric CT images along with their segmentation masks for longevity prediction. In \cite{Liu:2017} an architecture called XmasNet is provided that can maximize the accuracy of prostate cancer diagnosis. This network consists of $4$ Convolutional layers, $2$ fully connected layers, $2$ pooling layers and one SoftMax layer for cancer prediction. The inputs to this network are 3D MRI images.

\vspace{.05in}
In summary, self-designed CNNs are developed by varying the depth of the network (number of the Convolutional and non-Convolutional layers); the order the layers are cascaded one after another; the type of the input to the network (e.g., single channel or different form of multi-channel), and/or; the layer whose output is treated as the $\DLR$ features.

\vspace{.1in}
\item[1.1.3.] \textbf{\textit{Multiple CNNs}}:
Beside using single standard or self-designed CNNs, some researchers have proposed to use multiple networks, which has the advantage of benefiting from multiple inputs, having various modalities, scales and angles as shown in Fig.~\ref{fig:angle}, or different architectures with different properties.

\begin{figure}[t!]
\centering
\includegraphics[width=0.5\textwidth]{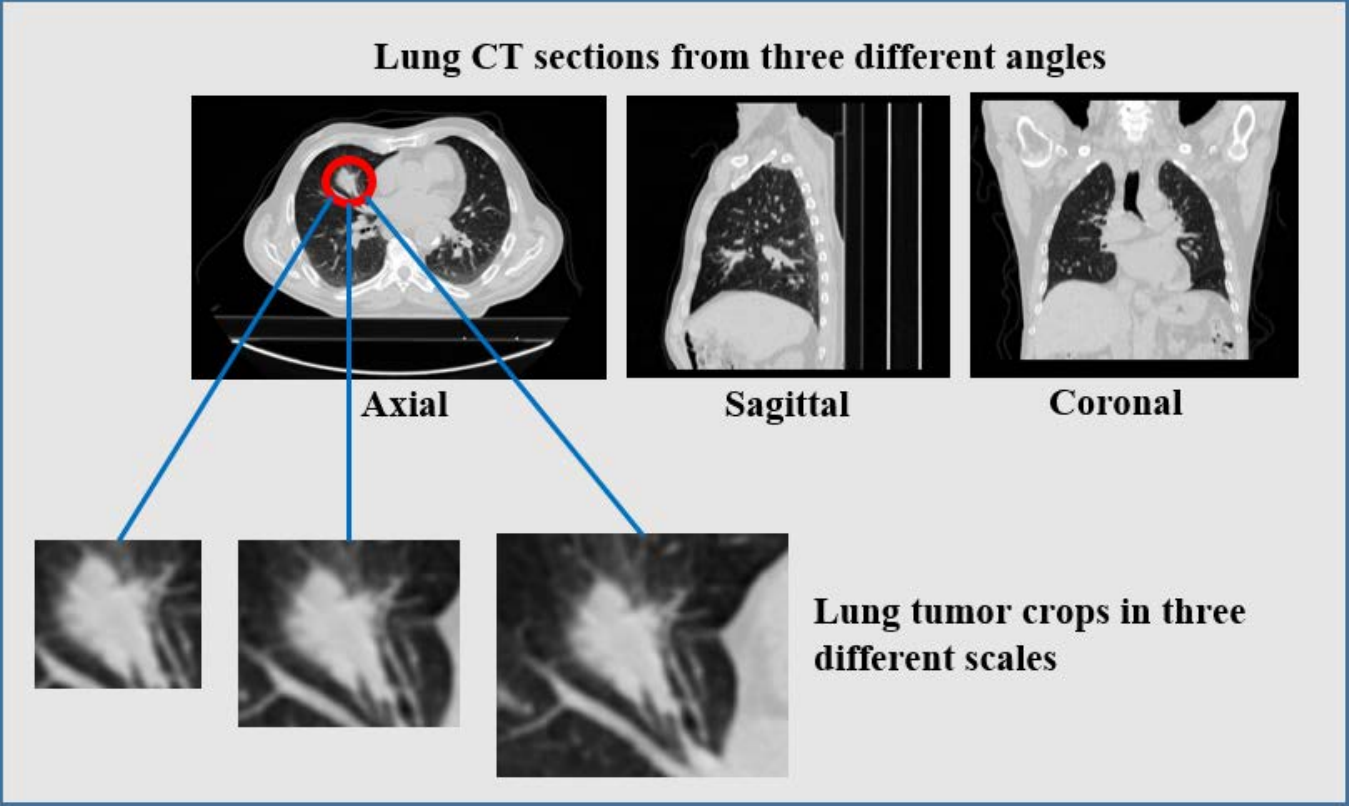}
 \centering
\caption{\small Different angles of lung CT scan along with tumor crops in three different scales.}
\label{fig:angle}
\end{figure}

\vspace{.05in}
``Scale'' is a significant factor to consider when designing the input structure. For example to distinguish tumors from vessels, a large enough region should be included in the input patch, while to differentiate between solid and non-slid tumors, the nodule region should be the main core of the patch. Having this in mind, Reference~\cite{Ciompi:2017} has designed a CNN architecture for lung tumor classification, where inputs are patches not only from different angles (sagittal, coronal, and axial) but also in different scales.
Following a similar path, Reference~\cite{Shen:2015} has also designed a multiple CNN architecture, where each CNN takes a lung tumor patch at a specific scale (illustrated in Fig.~\ref{fig:angle}) as input and generates the associated $\DLR$ features. Features extracted from all the CNNs are then concatenated and used for lung tumor malignancy prediction through a conventional classifier (SVM). The idea here is that segmenting the tumor regions is not always feasible. Furthermore using a tumor patch provides information on not only the tumor itself but also the surrounding tissues, and since tumor sizes can vary significantly among patients, using multi-scale patches instead of the single ones will improve the overall performance of the extracted $\DLR$ features. An interesting property of such multiple CNN architecture is that since the constituent CNNs share parameters, training can be performed in a reasonable time. Another benefit of using a multiple CNN architecture is that the network becomes robust to addition of small noise to the input.

\vspace{.05in}
Similar to the work in~\cite{Shen:2015}, Reference~\cite{KLiu:2017} has designed a CNN called ``Multi-view CNN", which uses 7 patches at different scales as inputs, with the difference that these patches are resized to have the same dimension, and therefore, a single CNN can be used instead of multiple CNNs. This work has also extended the binary lung tumor classification to a ternary classification to classify lung tumors as benign, primary malignant, and metastatic malignant. Furthermore, this article has adopted another validation approach called ``separability" besides the common terms such as accuracy and AUC (area under curve). Separability refers to the extend that different classes are distinguishable based on the learned features, and according to the aforementioned article, the proposed multi-view CNN has a higher Separability compared to a single scale CNN. In addition to that, as the layers go deeper, features with higher separability are learned.

\vspace{.05in}
The idea of using multi-scale image patches is further expanded in Reference~\cite{WShen:2017} through designing a novel CNN architecture called ``Multi-crop CNN", where instead of taking inputs in various scales, multi-scale features are extracted through parallel pooling layers, one of which applies pooling to a cropped version of the input from the previous layer. Features from multiple pooling layers are then concatenated and fed to the next layer. 3D lung CT images are inputs to this network, and since multiple CNNs are replaced with one single CNN, the training can be performed in a more time effective manner. Beside forecasting the lung tumor malignancy, this work has also predicted other attributes associated with tumor such as diameter, by replacing the final SoftMax layer with a regression one. It is worth mentioning that this network is not performing all the assigned tasks simultaneously. Instead they are performed one after another, which distinguishes this network from a multitask training one discussed in section~\ref{subsec:PtrRaw}.

\vspace{.05in}
Radiomics through multiple CNNs is further explored recently in~\cite{Liu:2018} for Alzheimer's disease diagnosis using MRI, where in the first stage several landmarks are detected based on the comparison between normal and abnormal brains. These landmarks are then used to extract patches (separately around each individual landmark), and consequently each CNN is trained taking patches corresponding to a specific landmark position as input. Final decision is made based on a majority voting among all the CNNs. Here, the idea behind using a multiple architecture is the fact that detecting Alzheimer's disease requires the examination of different regions of the brain.

\vspace{.05in}
In summary, multiple CNNs methods developed for $\DLR$ feature extraction are designed by either fusing the outputs of several CNNs which are trained based on a specific input, or multi-path layers are embedded within a single network to modify the output from previous layers differently.

\end{itemize}
\textit{One challenge shared among all the aforementioned CNN architectures is that they do not take the spatial information between objects into account. As an example, they may fail to consider the location of abnormality within the tissue as an indicator of its type.} The newly proposed deep architecture called CapsNets, described next, is introduced to overcome this drawback.

\vspace{.1in}
\noindent
\textit{\textbf{1.2. Capsule Networks}}:
\begin{figure*}[t!]
\centering
\includegraphics[width=1\textwidth]{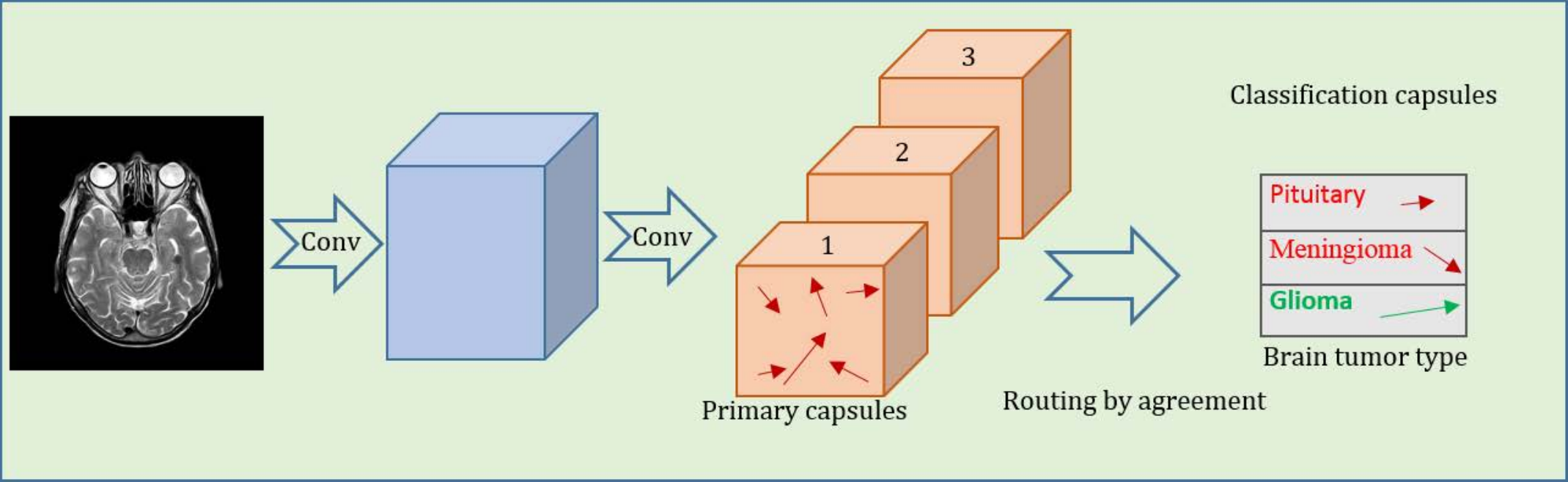}
 \centering
\caption{\small Capsule network architecture. A convolutional layer is used to form the primary capsules, and the decision is made based on the agreement among these capsules.}
\label{fig:caps}
\end{figure*}
Although CNNs are the state of the art in many medical and non-medical classification problems, they are subjected to several drawbacks including their low explainability and their negligence in preserving the spatial relationships between elements of the image leading to miss-classification. Besides, CNNs have low robustness to some types of transformation. Loss of spatial relation information, which is associated with the pooling layers, is resolved by the newly proposed Capsule networks (CapsNets)~\cite{Hinton:2017} consisting of both convolutional and capsule layers that can handle more types of transformation. These deep architectures have the ability to consider the relationships between the location of objects and tolerate more types of transformation, through their routing by agreement process, which dictates that an object will not be classified as a specific category unless the lower level elements of this object agree on the existence of that category. Another important property of CapsNets is that they can handle smaller datasets, which is typically the case in most medical areas. Here we explain the architecture of Capsule networks, as illustrated in Fig.~\ref{fig:caps}, and their routing by agreement process.

Capsules are group of neurons whose activity vectors consist of various instantiation parameters, and the length of the activity vectors represent the probability of a specific instance being present. Each Capsule in the primary capsules layer tries to predict the outcome of all the capsules in the next layer (parent capsules), however, these predictions are sent to the next layer with different coefficients, which is based on how much the actual output agrees with the prediction. This process of looking for agreement between capsules before coupling them is called routing by agreement and it is the most important property of capsule networks, making them consider spatial relations among objects, therefore, being robust to several types of transformations such as affine transformation and rotation. Defining $\u_i$ as the output of capsule $i$ in the primary capsules layers, the prediction for capsule $j$, $\u_{j|i}$ is calculated as follows, where $\W_{ij}$ is the weight matrix to be learned in back propagation

\begin{equation}
\centering
\u_{j|i}=\W_{ij}\u_i.
\end{equation}
Consequently $C_{ij}$, the coupling coefficient of capsules $i$ and $j$, is calculated based on the degree of conformation between these two capsules, based on the idea that if two vectors agree they will have a bigger inner product, and the output of parent capsule $j$, $\textbf{s}_j$, is estimated as follows

\begin{equation}
\centering
\s_j = C_{ij}\u_{j|i}.
\end{equation}
Finally, a non-linear function is applied to $\s_j$ to always keep its length equal or smaller than one as this length should represent probability of an object being present.

Below we further investigate the initial utilization of CapsNets in Radiomics~\cite{Parnian:ICIP18-p1}, for the first time. In \cite{Parnian:ICIP18-p1} we have explored various CapsNet architectures to select the one that maximizes the prediction accuracy resulting in a network that has fewer convolutional filters ($64$ filters) compared to the original Capsule network which has $256$ filters. This architecture consisting of one convolutional, one primary capsule, and a classification layer results in $86.56\%$ accuracy.

Furthermore, separate networks are trained based on two types of inputs: the original brain images and the segmented tumor regions, observing that CapsNet performs better when being fed with tumor masks, probably because the brain images have miscellanies backgrounds, distracting the network from extracting important distinguishing features. Nevertheless, Capsule network is of higher accuracy for both of the input types compared to the CNN, which has $78\%$ accuracy for tumor images. Several factors may have enabled the CapsNet to provide a better performance including its ability to handle small datasets, and being robust to transformations and rotations, resulted from the \textit{routing by agreement} process.

\vspace{.1in}
\noindent
\textit{\textbf{1.3. Recurrent Neural Networks}}:
Most of the deep network architectures need fixed-sized inputs, which makes them ineffective for Radiomics analysis of volumetric images (volume-level classification), i.e., when the whole volume is needed to be processed at once (such as tumor classification based on the 3D volume). In these scenarios, the RNNs can be adopted as they are capable of processing sequential data such as CT or MR slices, and they take both the present image slice and result of processing the previous ones as inputs. RNNs are also useful to monitor the medical images resulted from follow-up examinations (patient-level classification).

Since RNNs are associated with the vanishing gradient problem, a new type called long-short-term-memory (LSTM) is proposed which has the ability to decide what to store and what to forget. Although it seems that RNNs and LSTM are computationally more expensive than other architectures, their training time and cost is greatly reduced by using the same weights over the whole network \cite{Ravi:2017}. Use of LSTMs is explored in Reference~\cite{Azizi:2018} for prostate cancer benign and malignant classification based on sequences of ultrasound images, where it has been shown that the predictive accuracy of this sequential classification is higher than making decision based on independent single images.

This completes the overview of deep discriminative models with application to Radiomics. Next, we briefly review the generative models.

\vspace{.1in}
\noindent
\textbf{2. Generative Models}:
The objective of most of the deep generative models is to learn abstract yet rich features from the data distribution in order to generate new samples from the same distribution. What makes these models practical in Radiomics is the fact that the learned features are probably the best descriptors of the data, and thus have the potential to serve as Radiomics features and contribute to a consequent tasks such as tumor classification. Auto-encoder networks, deep belief networks, and deep Boltzmann machines are among the deep generative models that have been utilized in Radiomics works as outlined below:
\begin{itemize}
\item[2.1.] \textbf{\textit{Auto-Encoder Networks}}:
An auto-encoder network consists of two main components: An encoder which takes as input $\NS$ medical images denoted by $f^{(i)}$, for ($1 \leq i \leq \NS$), and converts each into a latent space $\phi (\W f\i + \b)$, i.e., Radiomics features.  The second component, the decoder, takes the latent space and tries to reconstruct the input image with the objective of minimizing the difference between the original input and the reconstructed one $\phi(\W^T\phi (\W f\i+\b)+\c)$~\cite{Kumar:2015} given by
\begin{equation}
\centering
\min _{\W, \b, \c}\sum_{i=1}^{\NS}||\phi\left(\W^T\phi \left(\W f\i+\b\right)+\c\right)-f\i||,
\end{equation}
where $\phi(\cdot)$ is the network's activation function; $\W$ denotes the weight matrix of the network used by both the encoder and the decoder; Term $\b$ denotes the encoder's bias vector; $\c$ is the decoder's bias vector, and; superscript $T$ denotes the transpose operator.
The reason that the encoded variables can be treated as Radiomics features is that they are the most important representatives of the input image that can be used to reproduce it. Although an auto-encoder can be trained completely in an end-to-end manner, to begin training with good initial weights and thus avoid the vanishing gradient problem, one can first train layers one by one, and use the obtained weights as the auto-encoder starting point \cite{Ravi:2017}. Depending on the application, Auto-encoders have several extensions including:
\begin{itemize}
\item[2.1.1.] \textbf{Denoising Auto-Encoders (DAEs)}: To make auto-encoders capture more robust features of the input, one common strategy is to add some noise to the input. This kind of auto-encoder is called a denoising auto-encoder (DAE) \cite{Ravi:2017}. Reference~\cite{Kim:2016} has adopted DAE for extracting Radiomics features that are fed to an SVM to classify lung tumors as benign or malignant. Reference~\cite{Sun:2017} has also adopted a five layer denoising auto-encoder which takes the corrupted lung images as inputs and tries to recover the original image. In particular, $400$  Features extracted by the encoder part of this network are treated as Radiomics to train another neural network for lung cancer classification (identify the type of the tumor such as benign or malignant).
\item[2.1.2.] \textbf{Convolutional Auto-Encoders (CAEs)}: This type of auto-encoders  are specially useful for Radiomics (image type inputs) as the spatial correlations are taken into account. In these networks, nodes share weights in a local neighborhood~\cite{Ravi:2017}. A CAE with $5$ Convolutional layer is adopted in Reference~\cite{Echaniz:2017} for lung cancer diagnosis (identify the presence of cancer).
\end{itemize}
There are two common strategies to leverage Auto-encoders in Radiomics:
\begin{itemize}
\item The first and most frequent approach is to directly use the extracted features to train a classifier. For instance, \cite{Kumar:2015} has extracted Radiomics features using a $5$ layer auto-encoder, which receives the segmented region of interest as the input. These features go through a binary  decision tree in the next step to produce the output which is the classified lung nodule in this case.
\item Auto-encoders can also serve as a pre-training stage to make the network extract representative features before trying to perform the actual classification. For instance, Reference \cite{Cheng:2016} has first trained a DAE based on resized (down-sampled images to facilitate training) lung CT patches. In the next stage, a classification layer is added to the network and the whole network is re-trained taking both resized images and the resizing ratio as inputs.
\end{itemize}
%
\item[2.2.] \textbf{\textit{Deep Belief Networks (DBNs)}}:
DBNs are stack of Restricted Boltzmann Machines (RBMs) on top of each other where the RBM is an unsupervised two layer stochastic neural network that can model probabilistic dependencies with the objective of minimizing the reconstruction error.  More importantly RBM is a bipartite graph allowing value propagation in both directions.
Although DBNs are composition of RBMs, only the top two layers have undirected relations. DBNs are first trained in a greedy fashion meaning that RBM sub-networks are trained individually followed by a fine-tuning phase~\cite{Ravi:2017}.  In Reference~\cite{Sun:2017}, a DBN consisting of $4$ hidden layers is designed with the goal of extracting the $\DLR$ from the top layer which has $1600$ nodes. This last layer is connected to an external neural network to classify lung nodules. Besides, to have multi-channel input (original image, segmented tumor, and gradient image), these channels are concatenated vector wise before being fed to the network.

\item[2.3.] \textit{\textbf{Deep Boltzmann Machine (DBMs)}}:
DBMs are also based on RBMs, but they differ from DBNs in the sense that DBMs include undirected relations between all layers which makes them computationally ineffective, though they are trained in a layer wise manner \cite{Ravi:2017}. Due to the two-way relations, however, RBMs can capture complicated patterns from the data~\cite{Suk:2014}. DBMs are adopted in~\cite{Suk:2014} for Alzheimer’s disease diagnosis. In this work, a classification layer is added to the last layer of the DBM allowing to extract not only hierarchical (generative) but also discriminative features.
\end{itemize}
This completes an overview of different deep discriminative and generative models used within the Radiomics workflow. Next, we consider a critical drawback of such architectures, i.e., acting as a black-box.
\subsection{Explainability of Deep Learning-based Radiomics}
Explainability of deep networks refers to revealing an insight of what has made the model to come into a specific decision, helping with not only improving the model by knowing what exactly is going on in the network, but also detecting the failure points of the model. No matter how powerful $\DLR$ are, they will not be utilized by physicians, unless they can be interpreted and related to the image landmarks used by the experts. Besides, not even a single mistake is allowed in medical decisions as it may lead to a irreparable loss or injury, and having an explanation of the logic behind the outcome of the deep net is the key to prevent such disasters. This subsection will present an overview on recently developed techniques to increase the explainability of deep Radiomics.

One simple approach to ensure the accuracy of the automatic prediction, is to double-check the results with an expert. For instance, Reference~\cite{Oakden-Rayner:2017}, which has used a CNN for longevity prediction using CT images, has reviewed the outcomes with experts leading to the fact that people predicted with longer lives are indeed healthier. However, this approach is time consuming and needs complete supervision and investigation, which is in conflict with the concept of automatizing and personalized treatment, which is the whole point of Radiomics. Therefore, nowadays several criteria are being presented to reduce the time and complexity of explaining deep Radiomics. One of these approaches is ``feature visualization'' which tries to gain knowledge on the network behavior by visualizing what kinds of features the network is looking for. This technique can be applied to different layers of the model. For example, to visualize the first layer features, the associated filters are applied to the input and the resulting feature maps are presented. However, as the last layer is the most responsible one in the network's output, paying attention to the features learned in this layer is more informative. For instance, Reference~\cite{Sun:2017} has visualized the final weights of a DBN, showing that the network is looking for meaningful features such as curvity. Nevertheless, these features are not as meaningful as they are for simple image recognition tasks as clinicians themselves are sometimes unsure about the distinctive properties of the images.

\begin{figure*}[t!]
\centering
\includegraphics[width=0.7\textwidth]{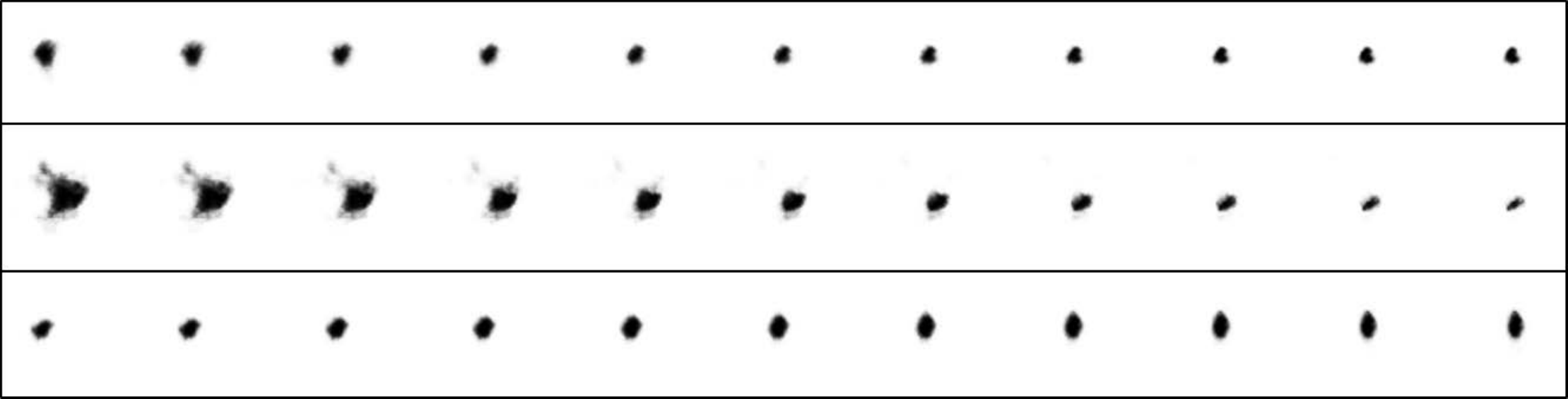}
\centering
\caption{\small Effect of tweaking the final feature vector on the reconstructed brain tumor image. Each row corresponds to a single feature which is tweaked 11 times.}
\label{fig:tweak}
\end{figure*}
One other method to provide the user with an explanation on the decision made by a deep architecture is called ``sensitivity analysis'' referring to generating a heat-map highlighting the image regions responsible for the output~\cite{Jamaludin:2016}. In the heat-map, the brighter areas are the ones that have influenced the prediction. This can be achieved by determining and measuring the effect of changing each individual input pixel on the output. In a CNN this effect can be estimated by determining the weight associated with each input pixel through back propagation. This approach can discover the cause for the prediction \cite{Jamaludin:2016}, however, the drawback of this approach is that not all the detected pixels through the heat-map are necessarily the ones leading to the specific decision, and besides, as the depth and complexity of the deep net increases, it becomes more difficult to measure the contribution of each individual pixel on the output.

A third proposed approach to understand the learned features is to project the high-dimensional feature space from the deep network to a bi-dimensional plane. Reference~\cite{Ciompi:2017} has adopted this strategy by using t-Distributed Stochastic Neighbor Embedding (t-SNE) algorithm to visualize the features learned by a CNN for lung tumor classification. The resulted plane presents clearly defined clusters of lung tumors, which shows that the networks has successfully learned discriminating features.  However, although this method can verify the accuracy of the network, it does not provide information on the exact reason behind making the decision.

The interpretability of meaningless weights is improved in the newly proposed Capsule networks through reconstructing the input image based on the features learned by the network. CapsNet includes a set of fully connected layers that take the final calculated features, based on which the final classification is made, as inputs, and reproduce the original image with the objective of minimizing the difference between the original and the reconstructed image. This objective function is added to the classification loss with a smaller weight not to distract network from extracting discriminative features. If the trained CapsNets is not only of high accuracy but also capable of resembling the input image, it has been successful in extracting representative features. Besides, visualizing these features provides insight on the explainability of the model. Interestingly, CapsNets are equipped with a powerful feature visualization technique through their input reconstruction part, which works as follows:
\begin{enumerate}
\item[1.] If CapsNet is in the training phase, the feature vector associated with the true class label is selected, otherwise, the one with the higher probability is used.
\item[2.] The selected feature vector is tweaked, meaning that small numbers are added to or subtracted from the feature values leading to a slightly changed new feature vector.
\item[3] The new feature vector is fed to the reconstruction part and the input image is reproduced. However this reconstructed image is not supposed to exactly resemble the input image as it is generated using the tweaked features not the actual ones learned by the network.
\item[4.] By repeating the process of tweaking and reconstructing process over and over again, one can understand what features are learned by observing the influence of changing them on the generated images.
\end{enumerate}
This strategy is adopted  for explaining the output of a CapsNet trained to classify brain tumors, where it is shown that the network is probably making the decision based on features such as size and deformation. Fig.~\ref{fig:tweak} shows the reconstructed brain tumor images based on tweaked feature vectors helping to gain insight on the nature of features.

This completes our discussion on deep learning-based Radiomics. Next, we consider hybrid solutions in which  hand-crafted and deep Radiomics are jointly used.
\begin{table*}[t!]
\centering
\small
\caption{\small A Comparison between hand-Crafted and Deep Radiomics.}
\vspace{-.1in}
\begin{tabularx}{\linewidth}{XX}
\arrayrulecolor{LightCyan}\hline
\rowcolor{LightCyan}
\textbf{~~~~~~~~~~~~~~~Hand-Crafted Radiomics ($\HCR$)} & \textbf{~~~~~~~~~~~~~~~~~~~~~~Deep Radiomics ($\DLR$)} \\
\arrayrulecolor{LightCyan}\hline
 Needs a prior knowledge on types of features to extract.&Can learn features on its own and without human intervention.\\
\\
Features are typically extracted from the segmented ROI. & does not necessarily require a segmented input.\\
\\
It is generally followed by a feature selection algorithm. & Feature selection is rarely performed.\\
\\
As features are defined independent from the data, does not require big datasets. & Requires huge datasets, since it has to learn features from the data.\\
\\
Processing time is not normally significant. & Can have high computational cost depending on the architecture and size of the dataset.\\
\\
Since features are pre-designed, they are tangible. & The logic behind the features and decisions is still a black box.\\
\\
\arrayrulecolor{LightCyan}\midrule
\end{tabularx}
\label{tab:comp}
\end{table*}
%
\section{Hybrid solutions to Radiomics} \label{sec:HybSol}

\begin{figure}[th]
\centering
\includegraphics[width=0.5\textwidth]{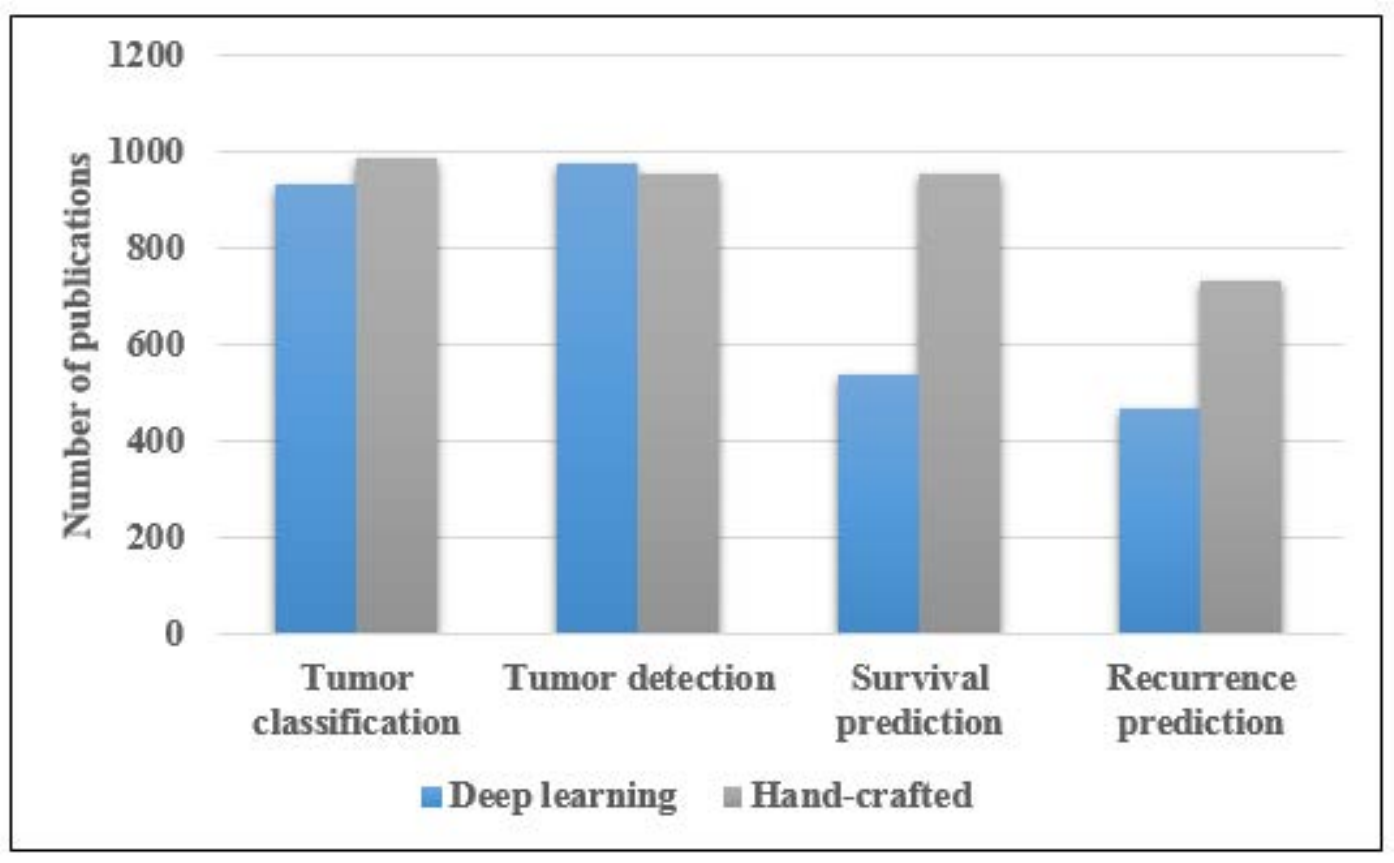}
\centering
\caption{\small Differences between the popularity of Hand-crafted and Deep learning-based Radiomics in four distinct applications, based on data from Google Scholar.}
\label{fig:DeHa}
\end{figure}

To summarize our findings on $\HCR$ and $\DLR$ features, tables~\ref{tab:comp}(a) and (b) provide different comparisons between these two categories from various perspectives. Fig.~\ref{fig:DeHa} shows the number of publications in four distinct applications of Radiomics, based on either hand-crafted or deep learning techniques. As it can be inferred from this figure, although deep learning techniques are much newer methods, compared to hand-crafted ones, these two techniques are on a par with each other, in tumor classification and detection. However, the number of publications on deep learning methods is relatively less than the number of publications on hand-crafted methods, in survival and recurrence prediction, possibly because the domain knowledge, which is not accessible in deep learning-based Radiomics, is more important in survival and recurrence prediction. This calls for techniques that can utilize both the advantages of deep learning (not requiring segmentation, feature selection, and human intervention), and the domain knowledge available in hand-crafted approaches.  Furthermore, in scenarios where neither of the above two mentioned categories are capable of providing informative Radiomics features with high predictive capacity, one can resort to hybrid strategies. Here, potential hybrid solutions to Radiomics~\cite{Emaminejad:2016} are reviewed from different points of view including combination of Radiomics with other data sources and combination of $\HCR$ and $\DLR$ features.

\subsection{Combination of Radiomics and Other Data Sources}
Physicians, normally, do not rely on a single input for their diagnosis of diseases and disorders.  To come into a conclusive decision, inputs from different sources are compared and combined including Radiomics (image bio-markers from different imaging modalities); Blood bio-markers; Clinical outcomes; Pathology, and; Genomics results~\cite{Lambin:2012}. In Information Post~\ref{infoPost1}, we have provided an overview of various Radiomics' imaging modalities and data sources which are typically combined with Radiomics features. Below, we discuss two different ways to fuse/combine Radiomics with other available resources of information along with the rationales and potentials behind such combinations:

\begin{itemize}
\item[i.] \textit{\textbf{Extracting Radiomics from Different Imaging Modalities}}: As stated previously, Radiomics can be extracted from different imaging modalities each of which can only capture/provide particular information on tissues'  properties. For instance, although the CT scan is among the most common and informative imaging modalities allowing to observe the body internal organs, the CT can not provide information on body function and metabolism. This type of information is available through PET scan, which calls for studying the effect of combining Radiomics extracted from different modalities. For example, in~\cite{Oikonomou:2018} Radiomics features are extracted from both CT and PET images, and the concatenated feature vector is fed to a classifier for lung cancer survival prediction, resulting in a higher accuracy compared to each modality separately.

\vspace{.05in}
Combining different imaging modalities is also tested on brain tumor classification in~\cite{Li:2017}. Since MRI can output different images varying mostly  in terms of their contrast and relaxation parameters, these images can be fused to provide complementary information. Based on~\cite{Li:2017}, extracting Radiomics from this combination of MRIs outperforms the single modal classifier for brain tumor classification.

\item[ii.] \textit{\textbf{Integration of Extracted Radiomics with Other Data Sources}}: Radiomics features are combined with other resources only after the extraction process. The best descriptive or predictive models in the field of Radiomics are the ones that utilize not only imaging bio-markers, but also other information such as Genomics patterns and tumor histology~\cite{Gillies:2016}.
In~\cite{Aerts:2014}, it is reported that combining Radiomics features with lung cancer staging information, which is obtained based on the tumor location and dispersion, can improve the prognostic performance of Radiomics alone or staging alone. In other words, Radiomics can provide a complementary information for lung cancer prognosis~\cite{Kumar:2012}. It is also shown that combining Radiomics with other prognostic markers in head-and-neck cancer leads to a more inclusive decision. Combining Radiomics with clinical data is further investigated in~\cite{Lao:2017} for brain cancer survival prediction. The interesting output of this work is a nomogram based on both Radiomics and clinical risk factors such as age that can be used to visually calculate survival probability.
Developing such a nomogram is further investigated in references \cite{Peng:2018} and \cite{Shen:2018} for prediction of Hepatitis B virus and lymph node metastasis, respectively.

In brief, the first step to build a Radiomics-based nomogram, is calculating a linear combination of selected Radiomics features based on a logistic regression, which results in a Radiomics score to exploit further for the desired prediction task. Consequently, by training a multivariate logistic regression, Radiomics score is fused with other influential factors to make the final prediction. Fig.~\ref{fig:nom} presents the nomogram introduced in \cite{Shen:2018} along with an example illustrating how the lymph node metastasis prediction is made.

\begin{figure*}[t!]
\centering
\includegraphics[width=0.7\textwidth]{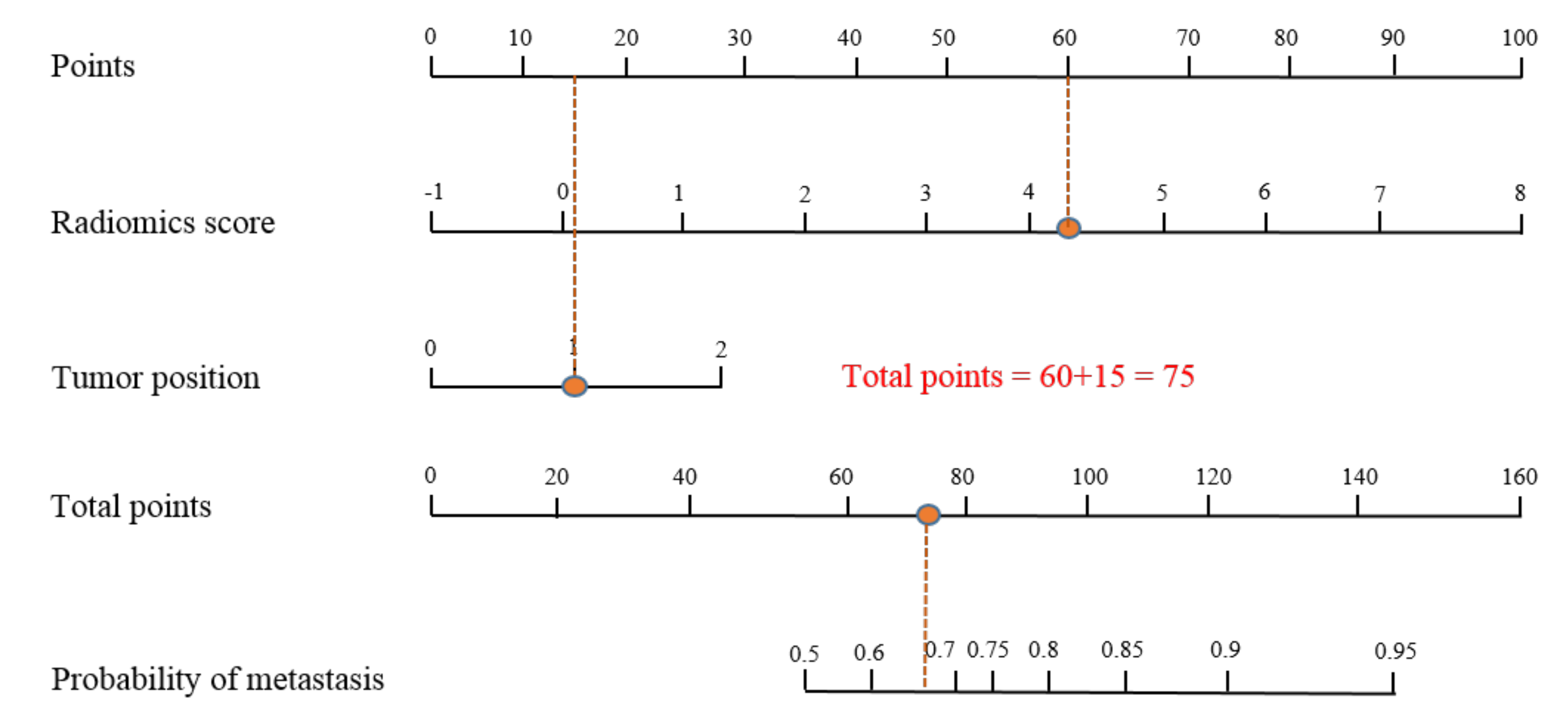}
\centering
\caption{\small Radiomics-based nomogram to predict lymph node metastasis. Tumor position is considered as extra information to assist with making a more reliable prediction.}
\label{fig:nom}
\end{figure*}
\vspace{.05in}

Reference~\cite{Emaminejad:2016} has adopted a fusion approach based on both Radiomics and Genomics bio-markers for predicting the recurrence risk of lung cancer. In this study, $35$ hand-crafted features are extracted from segmented lung CT images and reduced to $8$ after a feature selection phase. These features are then used to train a Naive Bayesian network. The same classifier is also trained using two Genomics bio-markers, and the outputs of two classifiers are fused through a simple averaging strategy. Results demonstrate that the combination of classifiers not only leads to higher prediction accuracy compared to individual ones, but also resembles the Kaplan-Meier plot of survival more precisely.
\end{itemize}

\subsection{Fusion of $\HCR$ with $\DLR$ (i.e., Engineered Features Coupled with Deep Features)}
\begin{figure*}[t!]
\centering
\includegraphics[width=0.7\textwidth]{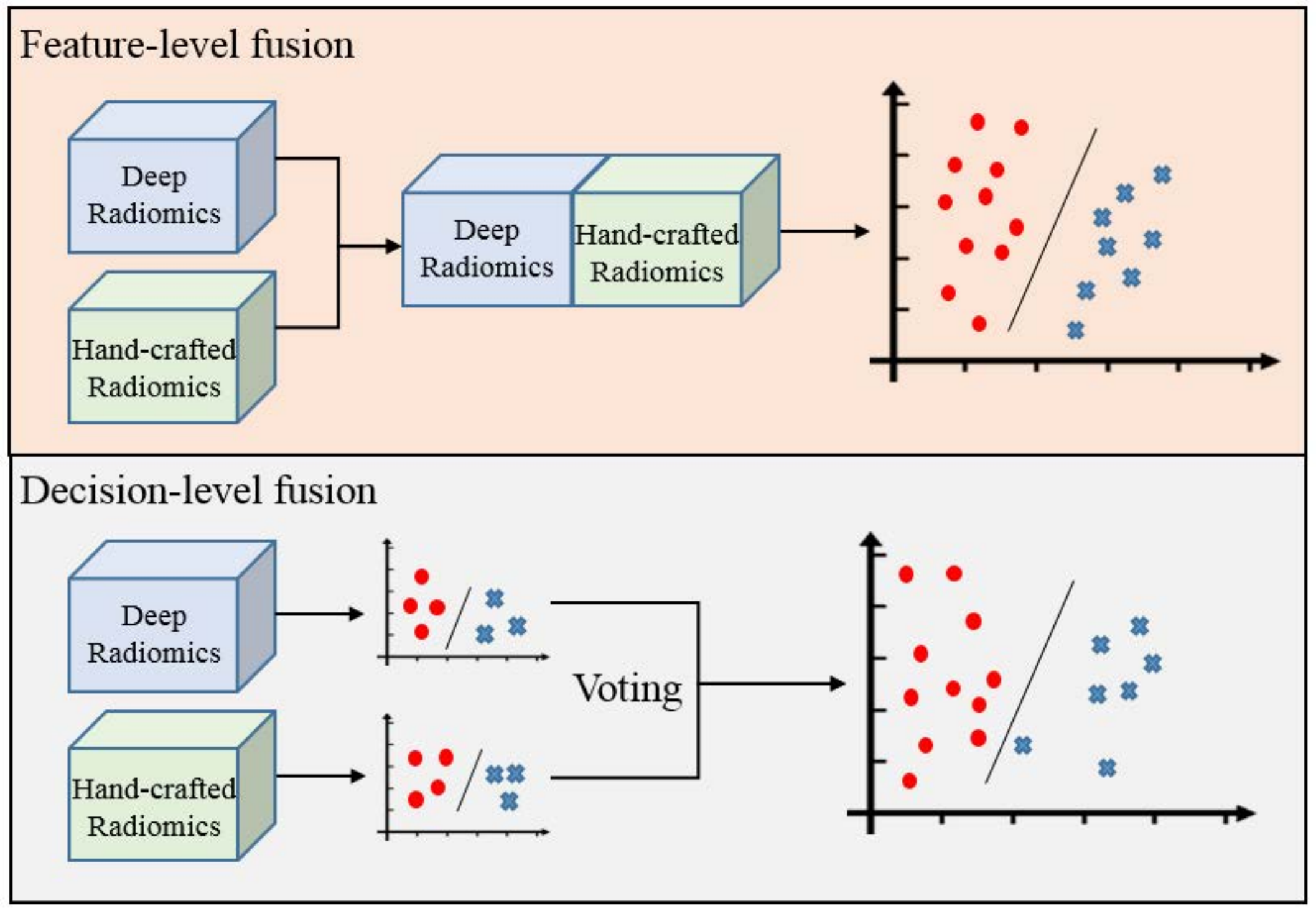}
\centering
\caption{\small Combining deep and hand-crafted Radiomics through feature-level or decision-level fusion.}
\label{fig:Com}
\end{figure*}
As mentioned in Table~\ref{tab:comp}, engineered (hand-crafted) and deep Radiomics both have their own advantages and disadvantages. As a result, combining these features has the promise of benefiting from both domains and incorporating different types of features~\cite{Fu:2017} potentially results in significantly improved performance.  As shown in Fig.~\ref{fig:Com}, the following two categories of data fusion have been used in Radiomics most recently:

\vspace{.1in}
\noindent
\textbf{1. Decision-level Fusion}:
One common approach to combine $\HCR$ with $\DLR$ is to first use them separately to train separate classifiers and then adopt a kind of voting between the outputs to make the final decision. The voting or fusion approaches in Radiomics include:
\begin{enumerate}
\item[(i)] \textit{\textbf{Soft Voting}}, which is combining the probability outputs, for instance, through a simple averaging.  Soft voting is adopted in~\cite{Li:2016}, where two individual SVMs are trained on hand-crafted and deep Radiomics features (extracted using a pre-trained CNN), and consequently breast cancer prediction is performed based on averaging the output probabilities. Results of this article shows that the combined features are associated with higher prediction accuracy. Fusion of separately trained classifiers through soft-voting based on deep and hand-crafted Radiomics is also examined in~\cite{Antropova:2017}  for breast cancer classification, where it has been shown that the combined SVM model outperforms individual classifiers in term of accuracy for mammogram, ultrasound, and MRI images.
\item[(ii)] \textit{\textbf{Hard Voting}}, which is combining outputs, for example, through a majority vote.
\item[(iii)] \textit{\textbf{Adaptive Voting}}, where a weight of importance for each model ($\HCR$ and $\DLR$) is learned for example using a separate  neural network. In Reference~\cite{Xie:2017}, a different kind of voting is adopted for lung cancer classification. This voting is based on the idea that not all the classifiers contribute equally to the final decision, and contribution weights are parameters that should be optimized through a learning process. The aforementioned article has first trained a CNN and several traditional classifiers such as SVM and logistic regression to independently predict the type (benign or malignant) of lung cancer. Predictions are consequently utilized to train a second-stage classifier to generate the final outcome. Any classifier such as SVM and NN can be used as the second-stage classifier.
\end{enumerate}
%

\vspace{.1in}
\noindent
\textbf{2. Feature-level Fusion}:
Second widely used approach to combine deep and hand-crafted Radiomics is to first concatenate the feature vectors and then feed them to a classifier, referred to as feature-level fusion~\cite{Kim:2016}.  Reference~\cite{Hawkins:2016} has shown that this combination lead to the highest performance in lung cancer survival prediction using Random Forest and Naive Bayes classifiers. The efficiency of this approach is also verified in~\cite{Fu:2017} for lung tumor detection. Although mixing deep and hand-crafted Radiomics has several advantages such as ensuring the heterogeneity of the features, it may cause over-fitting as the number of training data is relatively less than the number of features. Therefore, Reference~\cite{Lao:2017} has examined this large set of features in terms of stability, informativeness, and redundancy leading to a dramatic dimension reduction and increase in the accuracy of brain cancer survival prediction. To further reduce the number of features, a Cox regression is adopted that can determine the impact of features on survival, and as a result, those with small weights can be removed as effectless.

Reference~\cite{Chen:2017} has leveraged the idea of concatenating deep and hand-crafted features for lung tumor attributes (such as spiculation, sphericity and malignancy) scoring through a multi-task learning framework. For extracting deep features, $9$ CNNs corresponding to each of the $9$ task at hand, and a $3$ layer DAE are trained, where each CNN generates $192$ Radiomics features extracted from the last fully connected layer before the SoftMax layer, and DAE results in $100$ features. Deep features are further combined with hand-crafted features consisting of Haar and Histogram of oriented gradients (HoG) features, and the resulting vector is used as input to a multi-task linear regression model, which can consider the inter-task relations, in order to calculate the score of each of the $9$ lung cancer attributes. This completes our discussion on hybrid Radiomics.

\section{Challenges, Open Problems, and Opportunities} \label{sec:COO}
In this section, we will focus on the limitations of processing techniques  unique in nature to the Radiomics, and then introduce open problems and signal processing opportunities as outlined in the following subsections.

\subsection{Challenges of Hand-Crafted Radiomics}\label{Sec:CHR}
Extraction of hand-crafted features in Radiomics is more problematic in comparison to multimedia domains as, typically, very limited distinct visual variation exists to differentiate, for example, cancerous tissues. To address this issue, the common approach in Radiomics is to first extract hundreds and hundreds of different low level and high level features without devising a systematic mechanism and taking into account the end goal. Then, to resolve the resulting curse of dimensionality, simple and basic reduction techniques (e.g., basic principal component analysis (PCA)) are used. Another major issue  with existing hand-crafted features in Radiomics  (opportunity for signal processing researchers) is that they are extracted without using the information that can be obtained from other sources of data such as gene expressions and clinical data, which further limits their discriminative abilities for cancer prediction.

More importantly, most of the hand-crafted Radiomics require the segmented ROI. Providing annotations might not be a significant problem in other multi-media domains, as this project can be easily crowd sourced. However, when it comes to Radiomics, only experts have the ability to provide segmentations,  which is both time and cost ineffective.

\subsection{Challenges of Deep Radiomics}\label{Sec:CDR}
Although deep Radiomics has several advantages including its generalization capability and its independence from the supervision of experts, it is also associated with some disadvantages such as its need for huge data sets, and its lack of robustness to some kinds of transformation. Besides, another important challenge with deep learning-based Radiomics is that there is still no strategy to choose the optimum architecture.

Establishing the requirements of an appropriate deep architecture for Radiomics is a main challenge in development of $\DLR$ and another venue for future works. In particular, performing sensitivity analysis is a critical step to explain the connection between the designed choices and the achieved results. It is essential to identify different conditions under which the results are obtained. In short, what is happening when a specific architecture is used? For example, consider the simple issue of choosing the size of the input image. The question here is if  one should provide the original image size or downsize the image and why? One simple answer could be that downsizing is applied for computational savings. Another intuitive answer could be that down sampling makes the data invariant to translation and rotation, i.e., when the image is down sampled, we take averages and as such the model will be less prone to outliers. The latter intuition could be a reasonable idea for natural images but is this the case for Radiomics-based features obtained from medical images?

Intuitively speaking,  a fundamental challenge and an open problem for development of $\DLR$ is that the models and solutions developed for natural images should be modified before being applied to medical images as the nature of these images are totally different (e.g., consider MRI images). As the nature of the input signal is different, parts of the model has to change but this issue has not yet been investigated systematically and thoroughly.

It is worth mentioning that while hand-crafted Radiomics normally need less amount of data, the number of images required for effective training of deep architectures, such as the CNNs, depends on the complexity of the underlying model. In other words, the amount of required training data increases when the number of trainable parameters increase. In the case of lung cancer diagnosis, e.g., the LIDC Dataset~\cite{lidc:2015} is widely used, which consists of 244,527 CT images of 1010 patients. Normally, training  complex deep models based on such dataset cannot be performed  in a timely fashion using common processors and one needs to resort to one or more GPU processors. One common approach to reduce the computational cost in deep learning methods is to crop the input image to include just the object of interest. Although this approach does not cause any information loss in non-medical areas, it is harmful for classifying medical images as size is an important discriminator feature of, for instance, normal and abnormal tissues. This problem is investigated in \cite{Sun:2016}, where it is shown that the miss-classified lung tumors based on features extracted from a DBN are 4\% larger than correctly classified ones, possibly because the cropped images do not contain size information.

On the other hand, although limited access to training data is a common problem in other multimedia domains, it becomes significantly more critical in Radiomics as typically access to patients data is subject to several ethical  regulations making it hard to collect the required amount of data for training purposes.

Finally, interpretability of deep Radiomics is of paramount importance as human lives are at stake, and without providing appropriate explainability, utilization of deep-Radiomics in clinical practice will remain limited. The relation between deep features and genetic patterns is also not established yet and requires further studies.

\subsection{Open Problems and Signal Processing Opportunities}\label{Sec:OPO}
Despite recent advancements in the field of Radiomics and increase of its potential clinical applications, there are still several open problems which require extensive investigations including:
\begin{enumerate}
\item[1.] Most Radiomics models need rich amounts of training images, however, due to strict privacy policies, medical images are usually hard to collect.
\item[2.] Even without considering the privacy issues, it is difficult to find the required amount of data with similar clinical characteristics (e.g., corresponding to the same cancer stage).
\item[3.] Radiomics analysis need ground truth which is scarce as labels can only be provided by clinical experts  (this is in contrary to other multimedia domains). This calls for development of weakly or semi-supervised solutions taking into account the specifics of the Radiomics domain.
\item[4.] Properties of medical images such as their contrast and resolution varies significantly from one institute to another (from one dataset to another), because each institute may use different types of scanners and/or use different technical parameters. Development of novel and innovative information fusion methodologies together with construction of unifying schemes are critical to compensate for lack of standardization in this field and produce a common theme for comparing the Radiomics results. Furthermore ground truth and annotations provided by different experts can vary significantly as experts, depending on their area of specialty (such as oncology and surgery), may consider and look for different details and landmarks in the image.
\item[5.] Unbalanced data refers to a problem where classes are not equal in a classification problem rendering the classifier biased toward the majority class. This is almost all of the time the case for Radiomics analysis as the number of positive classes (existence of disease) is typically smaller than the negative ones. Therefore, proper care is needed when working with medical data. Although several solutions, such as modifying the metric function to give more weight to minority class, are provided to deal with the aforementioned issue, it is still an unsolved problem that needs further investigations.
\item[6.] Dealing with image noise is another challenging problem, which is common in all multi-media domains, but it is more severe in Radiomics as there may be more unpredictable sources of variation in medical imaging. As an example, patient's breathing in the CT scanner can cause change of lung tumor location in consecutive slices bringing about difficulty in extracting stable Radiomics features. Therefore to achieve reliable personalized diagnosis and treatment, careful strategies should be developed to address the effects of these kinds of variations.

Furthermore, there are several factors, such as imaging environments, capabilities of the scanners~\cite{Trinh:2014} and other shortcomings of radiological images (e.g., radiations during acquisition, noisy acquisitions), that limit the resolution of the obtained medical images. For instance, the range of the captured frequencies is limited by the maximum sampling rate of the scanner, and increasing the rate, will increase the resolution, at the cost of an increased noise~\cite{Greenspan:2009}. Since access to high-quality images is necessary to achieve an early and accurate diagnosis/detection, there is an ongoing research on improving the quality of the medical images via development of advanced computational models to overcome the aforementioned shortcomings. One of such computational techniques is known as ``Super-Resolution~\cite{Dong:2014}'', aiming at reconstructing a high-resolution image, using several low-resolution instances. Deep learning networks, and CNNs in particular, are widely used in super-resolution problems, and so far have shown promising results~\cite{Dong:2014}.
\item[7.] The biggest challenge in combining various data sources (such as imaging and clinical) is that not all data is provided for all the patients. In other words, Radiomics analysis model should be equipped with the ability to work with sparse data~\cite{Gillies:2016}. Besides, The currently used fusion strategies within the Radiomics are still in their infancy and development of more rigorous fusion rules is necessary. For instance, feature-level fusion results in a vector and how to sort/combine the localized feature vectors is an open challenge. Giving the superiority of the initial results obtained from hybrid Radiomics, this issue becomes an urgent matter calling for advanced multiple-model solutions.
\end{enumerate}
%
\section{Conclusion} \label{sec:Conc}
During the past decades, medical imaging made significant advancements leading to the emergence of automatic techniques to extract information that are hidden to human eye. Nowadays, the extraction of quantitative or semi-quantitative features from medical images, referred to as Radiomics, can provide assistance in clinical care especially for cancer diagnosis/prognosis. There are several approaches to Radiomics including extracting hand-crafted features, deep features, and hybrid schemes. In this article, we have presented an integrated sketch on Radiomics by introducing practical application examples; Basic processing modules of the Radiomics, and; Supporting resources (e.g., image, clinical, and genomic data sources) utilized within the Radiomics pipeline, with the hope to facilitate further investigations and advancements in this field within signal processing community.

\bibliographystyle{IEEEbib}

\small
\vspace{.1in}
\noindent
\textbf{Parnian Afshar} (p\_afs@encs.concordia.ca) is a Ph.D. candidate at Concordia Institute for Information System Engineering (CIISE). Her research interests include signal processing, biometrics, image and video processing, pattern recognition, and machine learning. She has extensive research/publication record in medical image processing related areas.

\small
\vspace{.1in}
\noindent
\textbf{Arash Mohammadi} (arash.mohammadi@concordia.ca) is an Assistant Professor with Concordia Institute for Information System Engineering (CIISE), at Concordia University. He is the Director-Membership Services of IEEE Signal Processing Society (SPS), and was General co-chair of the Symposium on “Advanced Bio-Signal Processing and Machine Learning for Medical Cyber-Physical Systems,” under IEEE GlobalSIP 2018, and the lead Organizer of the 2018 IEEE SPS Video and Image Processing
(VIP) Cup. He also was the General Co-Chair of 2016 IEEE SPS Winter School on “Distributed Signal Processing for
Secure Cyber Physical Systems”. His research interests include statistical signal processing, biomedical signal processing, image/video processing, and machine learning.

\small
\vspace{.1in}
\noindent
\textbf{Konstantinos N. Plataniotis} (kostas@ece.utoronto.ca) Bell Canada Chair in Multimedia, is a Professor with the ECE Department at the University of Toronto.  He is a registered professional engineer in Ontario, Fellow of the IEEE and Fellow of the Engineering Institute of Canada. Dr. Plataniotis was the IEEE Signal Processing Society inaugural Vice President for Membership (2014-2016) and the General Co-Chair for the IEEE GlobalSIP 2017. He co-chaired the 2018 IEEE International Conference on Image Processing (ICIP 2018), October 7-10, 2018, Athens Greece, and co-chairs 2021 IEEE International Conference in Acoustics, Speech \& Signal Processing (ICASSP 2021), Toronto, Canada.

\small
\vspace{.1in}
\noindent
\textbf{Anastasia Oikonomou} (anastasia.oikonomou@sunnybrook.ca), is the head of the Cardiothoracic Imaging Division at Sunnybrook Health Science Centre; Site Director of the Cardiothoracic Imaging Fellowship program at University of Toronto, and an Assistant Professor with the Department of Medical Imaging at the University of Toronto. Her research interests include imaging of pulmonary malignancies and interstitial lung diseases, Radiomics and machine learning methods in imaging of pulmonary disease.

\small
\vspace{.1in}
\noindent
\textbf{Habib Benali} (habib.benali@concordia.ca) is a Canada Research Chair in ``Biomedical Imaging and Healthy Aging,'' is the Interim Scientific Director of the PERFORM Centre, and a Professor with the ECE Department at Concordia University. Dr. Benali was the director of Unit 678, Laboratory of Functional Imaging of the French National Institute of Health and Medical Research (INSERM) and Paris 6 University (UPMC). He was also the director of the International Laboratory of Neuroimaging
and Modelisation of the INSERM-UPMC and Montreal University. His research interests include biological signal processing, neurology and radiology, biomedical imaging, biostatistics, and bioinformatics.	
\printnomenclature[1.5in]

\end{document}